\PassOptionsToPackage{expansion=false}{microtype}
\PassOptionsToPackage{table,xcdraw}{xcolor}
\documentclass[]{bytedance_seed}

\usepackage[toc,page,header]{appendix}
\usepackage{microtype}
\usepackage{graphicx}
\usepackage{subcaption}
\usepackage{booktabs} 
\usepackage{float}
\usepackage{hyperref}
\usepackage{bm}

\usepackage{amsmath}
\usepackage{amssymb}
\usepackage{mathtools}
\usepackage{amsthm}
\theoremstyle{plain}

\theoremstyle{definition}

\theoremstyle{remark}

\usepackage{xspace}
\usepackage{wrapfig}
\usepackage{multirow}   
\usepackage{tabularx}      
\usepackage{colortbl}  
\usepackage[export]{adjustbox}

\renewcommand{\titlefont}{\fontsize{17}{19.5}\selectfont}
\renewcommand{\authorfont}{\fontsize{10.4}{12.2}\selectfont}
\renewcommand{\affiliationfont}{\fontsize{8.8}{10.2}\selectfont}
\renewcommand{\contributionfont}{\fontsize{8.4}{10}\selectfont}
\renewcommand{\abstractinfont}{\fontsize{9.4}{11.2}\selectfont}

\title{UMI-Bench 1.0: An Open and Reproducible\\
Real-World Benchmark for Tabletop Robotic\\
Manipulation with UMI Data}

\renewcommand{\authorlist}{%
{\authorfont\sffamily
Shi Jin\textsuperscript{1,2,\textdagger} \quad Yuntian Wang\textsuperscript{1,2,\textdagger} \quad Yuhui Duan\textsuperscript{1,2,\textdagger} \quad Di Wu\textsuperscript{1,2,\textdagger} \quad Gaoqi Dong\textsuperscript{2,\textdagger}\\[1.5pt]
Xiaohang Liu\textsuperscript{2,\textsection} \quad Xiaotong Li\textsuperscript{2,\textsection} \quad Hongfei Jia\textsuperscript{2,\textsection} \quad Zehao Zhang\textsuperscript{2,\textsection} \quad Tianyu Wang\textsuperscript{3}\\[1.5pt]
Zhongjie Jia\textsuperscript{4} \quad Yuanqi Yao\textsuperscript{7} \quad Chenjia Bai\textsuperscript{5} \quad Zhaxizhuoma\textsuperscript{4,6,2,\textdaggerdbl} \quad Siao Liu\textsuperscript{1,\textdaggerdbl}\\[1.5pt]
Nieqing Cao\textsuperscript{8} \quad Jin Wang\textsuperscript{1} \quad Chao Yu\textsuperscript{2} \quad Yan Ding\textsuperscript{2,3,8}}}

\renewcommand{\affiliationlist}{%
{\affiliationfont\fontfamily{bytesansmedium}\selectfont
\textsuperscript{1}Soochow University \quad \textsuperscript{2}Lumos Robotics \quad \textsuperscript{3}Fudan University \quad \textsuperscript{4}Shanghai Jiao Tong University\\[1pt]
\textsuperscript{5}Shanghai TeleAI \quad \textsuperscript{6}Shanghai AI Laboratory \quad \textsuperscript{7}INSAIT \quad \textsuperscript{8}Xi'an Jiaotong-Liverpool University}}

\renewcommand{\contributionlist}{%
{\contributionfont
\textsuperscript{\textdagger}Equal contribution \quad \textsuperscript{\textsection}Equal contribution \quad \textsuperscript{\textdaggerdbl}Corresponding authors}}

\correspondence{\email{zak-zxzm@sjtu.edu.cn}; \email{saliu@suda.edu.cn}}
\checkdata[Project Page]{\url{https://umibenchmark.github.io/}}
\checkdata[Keywords]{Robot Manipulation, Real-world Benchmark, UMI Data}

\abstract{%
Real-robot evaluation is essential for understanding whether learned manipulation policies can operate reliably outside curated demonstrations.
This need is particularly pressing for Universal Manipulation Interface (UMI)-style policies, whose performance depends on the coupling between wrist-view observations, action representation, data collection, and physical deployment.
Existing real-world benchmarks have made important progress, but they are not designed around this UMI data-to-deployment setting.
We present UMI-Bench 1.0, a local-first real-robot benchmark for standardized evaluation of UMI-style manipulation policies.
To the best of our knowledge, this is the first benchmark dedicated to real-world evaluation of UMI-based manipulation models.
UMI-Bench aligns data collection, scene reset, policy execution, result logging, and task-factor analysis within a unified protocol.
By making the full evaluation process reproducible and auditable, UMI-Bench provides a practical testbed for measuring how UMI-trained policies generalize to real physical manipulation.
}

\begin{document}
\maketitle

\begin{figure}[t]
    \centering
    \includegraphics[width=0.9\linewidth]{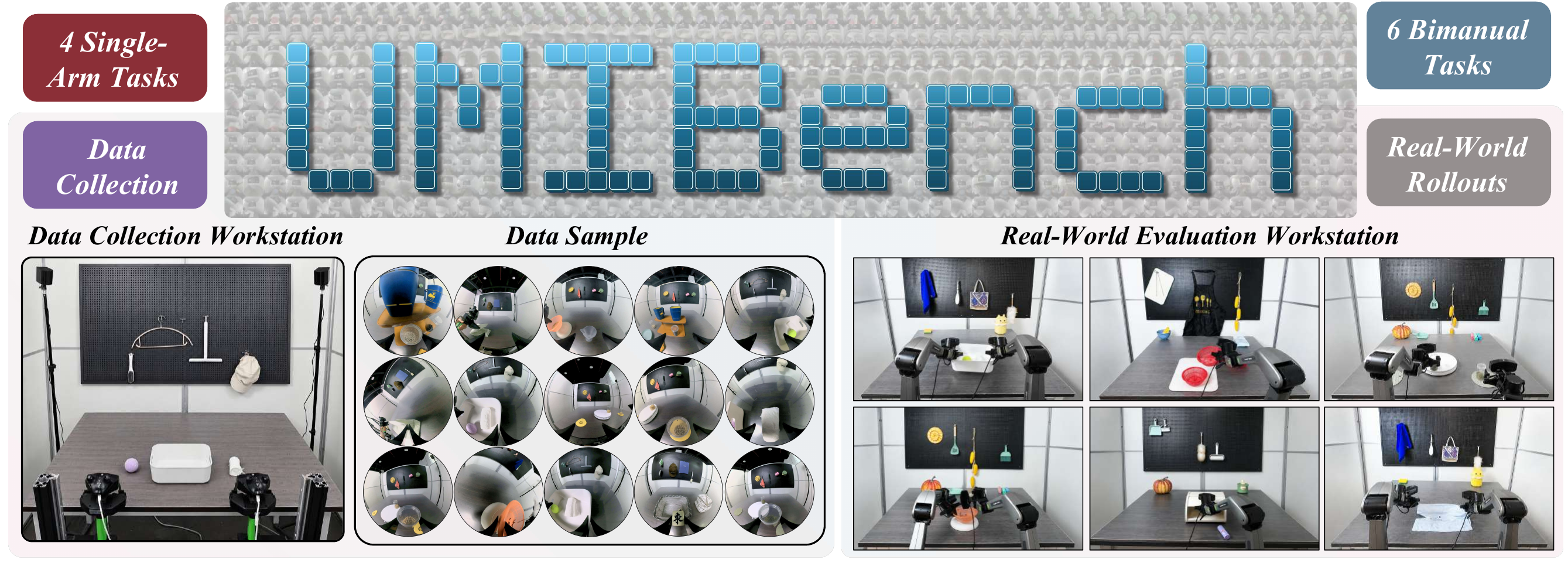}
    \caption{\textbf{Overview of UMI-Bench 1.0}
    The benchmark links standardized data collection workstations and demonstration samples with real-world evaluation workstations, covering four single-arm tasks and six bimanual tasks and producing real-world rollouts for reproducible policy evaluation.}
    \label{fig:overview}
\end{figure}

\section{Introduction}
\label{sec:intro}

Vision-language-action models have made rapid progress on robotic manipulation tasks~\cite{pi_0,pi_0_5,dreamzero,act,dp}.
As these models move toward real deployment, evaluation on physical robots becomes increasingly important.
Simulation and offline metrics provide useful development signals, but they cannot \emph{fully} capture contact dynamics, sensing artifacts, timing variation, and operator-induced differences in real manipulation.
Real-robot benchmarking is therefore necessary for measuring whether a policy can close the loop under physical execution noise.

Recent benchmarks have begun to address this problem.
RoboChallenge~\cite{robochallenge} studies large-scale real-robot evaluation and highlights the difficulty of making physical tests scalable and reproducible.
RoboArena~\cite{roboarena} and ManipArena~\cite{maniparena} further show the value of unified real-world protocols for evaluating generalist robot policies.
However, these benchmarks are not designed around UMI data, which has become an important form of real-world manipulation data for scalable robot learning.
Unlike conventional robot demonstrations, UMI data is organized around a deployment-oriented interface in which wrist-view observation, action representation, embodiment, scene reset, and policy execution are closely linked~\cite{umi,fastumi,fastumi100k}.
A benchmark for UMI-based policies should therefore evaluate not only task completion, but also whether the complete UMI data-to-deployment pipeline can be reproduced in the real world.

This missing alignment makes real-world comparison difficult.
When policies are tested with different hardware setups, camera placements, reset procedures, or action interfaces, performance differences can reflect evaluation variation rather than model capability.
When scene conditions are specified only informally, failure analysis cannot reliably separate object-level generalization, layout variation, spatial reasoning, or execution instability.
When data collection and evaluation are specified independently, the relationship between the training distribution and the test distribution remains unclear.

We introduce UMI-Bench 1.0 to address this gap.
UMI-Bench is a local-first real-robot benchmark for UMI-style wrist-view manipulation policies.
It standardizes the data-to-evaluation pipeline, including demonstration collection, episode specification, scene reset, policy execution, logging, scoring, and task-factor analysis.
The resulting benchmark is a reproducible evaluation protocol rather than a collection of isolated tabletop tasks.
Its first release instantiates this protocol on tabletop manipulation tasks with standardized demonstrations and real-world evaluation episodes.
UMI-Bench evaluates task success, partial progress, and subgoal completion under both seen and unseen conditions. 
To make generalization failures interpretable, unseen conditions are organized by task-defined variation factors, and tasks are annotated with capability-relevant properties such as spatial reasoning, grasp stability, placement precision, contact interaction, and multi-stage execution.

Our contributions are as follows.
First, we formulate UMI-Bench 1.0 as a real-robot benchmark for \textbf{UMI-style} manipulation policies.
Second, we standardize the local evaluation protocol to reduce variation in hardware setup, observations, scene reset, and result logging.
Third, we connect standardized UMI data collection with real-robot evaluation through a unified data schema, episode specification, and evaluation runner.
Fourth, we define task metrics and task-factor analysis for diagnosing policy performance under seen and unseen evaluation conditions.

\section{Related Work}
\label{sec:related}

\noindent\textbf{Robot evaluation benchmarks.}
RoboChallenge~\cite{robochallenge}, RoboArena~\cite{roboarena}, and ManipArena~\cite{maniparena} make real-machine evaluation more systematic through unified task suites and protocols for generalist robot policies.
Simulation benchmarks such as RLBench~\cite{rlbench}, LIBERO~\cite{libero}, ManiSkill~\cite{maniskill}, RoboCasa~\cite{robocase}, and VLABench~\cite{vlabench} provide scalable environments with precise control over initial states, object assets, and success conditions, though simulated evaluation cannot fully capture camera artifacts, calibration drift, frictional contact, and hardware-specific timing effects.
UMI-Bench is complementary to both: it borrows the idea of controlled evaluation splits but instantiates them in physical scenes, and focuses on reproducibility for UMI-trained policies by specifying how data are collected, how wrist-view observations are generated, how scenes are reset, and how evaluation logs relate to the training distribution.

\noindent\textbf{Real-world robot data and generalist policies.}
BridgeData V2~\cite{bridgedata_v2}, Open X-Embodiment~\cite{openx}, and DROID~\cite{droid} have scaled real-world manipulation data across platforms, embodiments, and environments.
At the model level, RT-1~\cite{rt1}, RT-2~\cite{rt2}, Octo~\cite{octo}, $\pi_0$~\cite{pi_0}, $\pi_{0.5}$~\cite{pi_0_5}, and DreamZero~\cite{dreamzero} show that scaling data and model capacity can produce increasingly general robot policies.
However, a large dataset or a strong generalist policy does not by itself define a fair real-robot evaluation protocol.
UMI-Bench therefore focuses on the missing link between standardized data production and standardized physical evaluation.

\noindent\textbf{UMI-style data collection.}
The Universal Manipulation Interface provides a practical route for collecting manipulation demonstrations with a deployment-oriented observation interface~\cite{umi}.
FastUMI extends this direction by improving scalability and hardware independence for UMI-style data collection~\cite{fastumi,fastumi100k}.
These works make it easier to collect useful real-world demonstrations, but they leave open an evaluation question: how should policies trained from such data be tested on real robots under a reproducible protocol.
UMI-Bench answers this question by treating collection, metadata, scene reset, action execution, and scoring as one coupled benchmark specification.


\begin{figure}[t]
    \centering
    \includegraphics[width=0.95\linewidth]{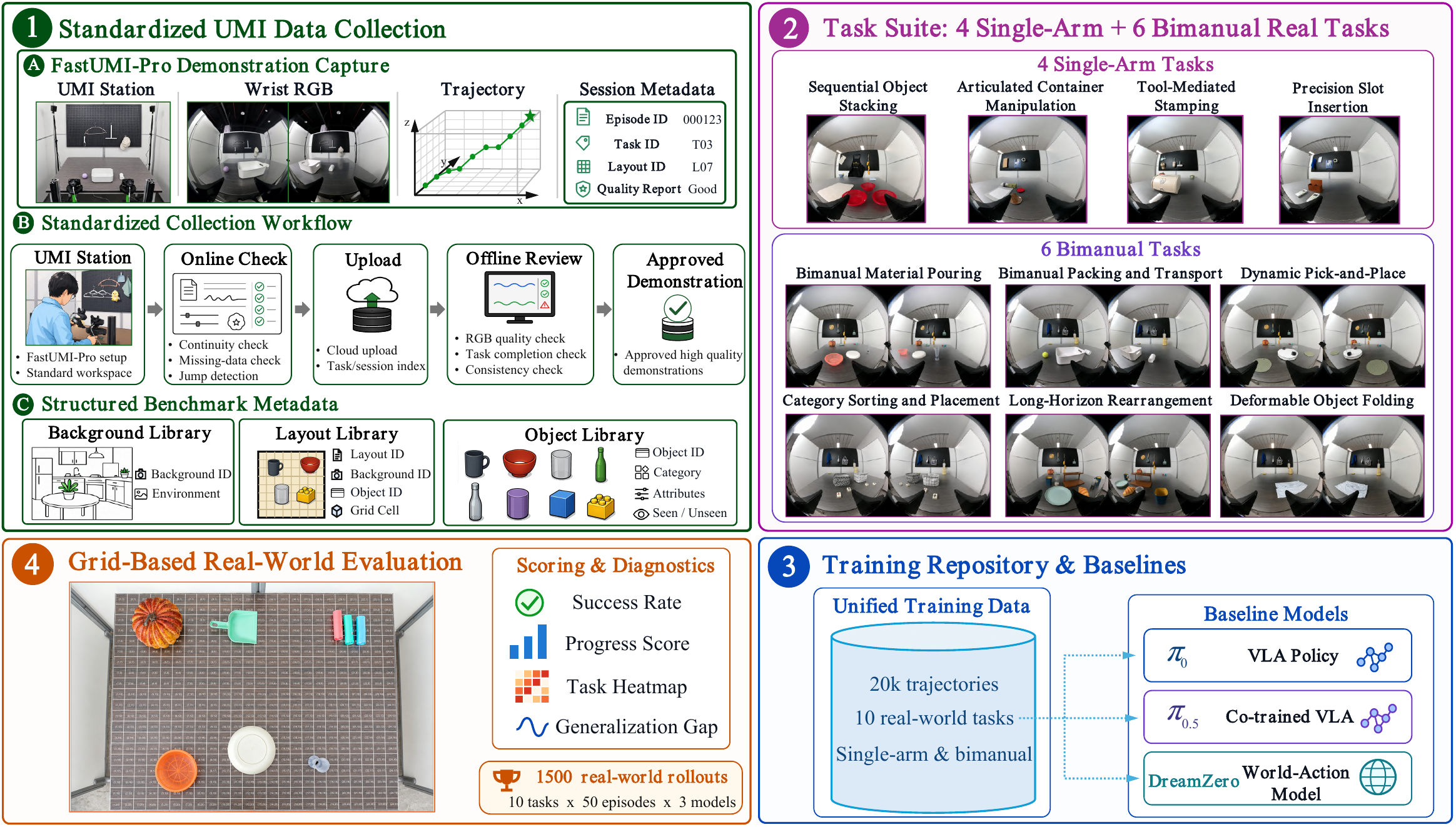}
    \caption{\textbf{Data-to-evaluation pipeline.}
    UMI-Bench connects standardized UMI data collection, structured benchmark metadata, a 10-task manipulation suite, unified training repositories, baseline policies, and real-world evaluation with success, progress, task-level, and generalization diagnostics.}
    \label{fig:pipeline}
\end{figure}

\section{UMI-Bench Benchmark}
\label{sec:benchmark}

This section specifies UMI-Bench 1.0 as an end-to-end benchmark, following the data-to-evaluation organization in Fig.~\ref{fig:pipeline}.
We first specify the benchmark design requirements and workstation interface, then describe standardized UMI data collection, the task suite and generalization factors, and the real-world evaluation protocol and diagnostics.

\subsection{Benchmark Design}
\label{sec:design}

UMI-Bench is designed around three requirements.
First, it must be aligned with UMI-style data.
Specifically, the observation interface, action space, data schema, and evaluation runner should be consistent with the assumptions used during UMI-style policy training.
Second, it must be reproducible.
The hardware configuration, tabletop workspace, object reset procedure, episode metadata, and scoring protocol should be specified in sufficient detail for another site to reconstruct the benchmark.
Third, it must provide diagnostic value.
The benchmark should reveal not only whether a policy succeeds, but also which task factors and distribution shifts contribute to failure.
This subsection focuses on the evaluation-side platform and protocol.
Subsection~\ref{sec:data} describes the standardized data collection pipeline used to construct the training repository.

\noindent\textbf{System Overview.}
The UMI-Bench pipeline is illustrated in Fig.~\ref{fig:pipeline}.
At a high level, standardized demonstrations are converted into a training repository, and trained checkpoints are evaluated on a local robot workstation using a shared episode specification.
On the evaluation side, the pipeline standardizes the hardware platform, wrist-view observation interface, action abstraction, scene reset protocol, rollout logging, and human scoring workflow.
To reduce avoidable hardware variation in the first release, this platform is instantiated as a standardized tabletop robot workstation, including the robot arm, gripper, wrist-camera mount, control interface, and safety configuration; detailed specifications are provided in Appendix~\ref{app:hardware}.
The runner records a rollout package containing the scene JSON, wrist-view video, trajectory logs, action logs, and runtime metadata. 
Human evaluators subsequently annotate task success and progress scores from the recorded rollout.

\noindent\textbf{Wrist-View Observation Interface.}
Policies are evaluated using wrist-view RGB observations, optionally with robot proprioception when required by the method.
Third-person cameras are not used as policy input.
We record wrist-view videos for auditing because they correspond to the actual observation stream seen by the policy; third-person videos may be collected only for debugging and qualitative analysis.

\noindent\textbf{Workspace and Scene Reset.}
The benchmark uses a fixed tabletop workspace with predefined robot placement, object regions, lighting conditions, and reset procedures.
The current workstation specification uses a 1.2 m by 1.0 m by 0.75 m tabletop with a matching acrylic grid board on the surface.
The grid contains human-readable 5 cm by 5 cm cells for object reset and metadata recording.
It is intended for operators, not for policies.
The grid is visible enough to support consistent reset, but is not assumed to be reliably decoded by the wrist camera.

Each evaluation episode is specified by a reset image and a structured scene JSON file.
The scene JSON records the task identifier, object identifier, category, appearance, material, tabletop marker position, pose, target region, and split.
This representation extends reference-image reset with explicit distribution metadata.
It allows the benchmark to report performance by object split, layout split, task factor, and subgoal.

\subsection{Standardized UMI Data Collection}
\label{sec:data}

This subsection describes how the training data are collected and organized.
It complements the evaluation platform in Subsection~\ref{sec:design} by specifying the data provenance, collection interface, and demonstration repository.

\noindent\textbf{Demonstration Management Interface.}
The web collection system provides the administrative layer for standardized data collection.
It manages task definitions, collector assignment, episode indexing, metadata entry, quality-control status, and dataset export.
Each demonstration is associated with a task prompt, object metadata, scene metadata, wrist-camera video, robot state, action chunks, timestamps, and quality flags.
This layer reduces ambiguity in data provenance and makes the collected demonstrations auditable.

\noindent\textbf{Table-Task Collection Workstation.}
The table-task collection workstation provides the physical layer for standardized collection.
It fixes the tabletop, robot placement, wrist camera, object regions, operator interface, and FastUMI Pro collection setup.
This standardization reduces variation across collectors and collection sessions.
It also makes the training distribution more directly comparable with the evaluation distribution.

\noindent\textbf{UMI-Style Collection Protocol.}
FastUMI Pro is used as part of the data collection toolchain.
The collection procedure consists of hardware checks, wrist-camera checks, time synchronization, calibration verification, task selection, scene reset, demonstration recording, failure annotation, quality control, and format conversion.
The main paper summarizes this workflow because it is part of the benchmark protocol.
Appendix~\ref{app:fastumi_pro} gives the detailed procedure, and Appendix~\ref{app:data_schema} summarizes the exported data schema.

\noindent\textbf{Demonstration Repository.}
The demonstration repository stores both training-ready data and raw collection metadata.
Each task entry contains demonstrations, prompt templates, object lists, collection settings, a scene metadata schema, an action-space specification, and a test split.
The first release contains 10 tasks with 1,600--3,000 demonstrations per task, resulting in approximately 20k demonstrations in total.

\subsection{Task Suite and Generalization Design}
\label{sec:tasks}

\noindent\textbf{Task Suite.}
It contains 10 real-world tabletop tasks: 4 single-arm tasks and 6 bimanual tasks.
It covers stacking, articulated-container manipulation, tool use, multi-object storage, material transfer, coordinated carrying, dynamic object grasping, category sorting, deformable-object manipulation, and long-horizon rearrangement.
The tasks are selected to expose real physical failure modes while remaining feasible for current policies.
Table~\ref{tab:task_suite} provides the task-level specification, and Appendix~\ref{app:task_cards} provides a variant-level coverage visualization with representative scene examples.
Task names in Table~\ref{tab:task_suite} denote manipulation task types, while the goals specify the concrete real-world instantiations used in UMI-Bench 1.0.
The task IDs in Table~\ref{tab:task_suite} are used in subsequent summary tables.

\begin{table}[t]
    \centering
    \caption{Task suite in UMI-Bench 1.0. Demonstration and real-world evaluation episode counts are listed per task.}
    \label{tab:task_suite}
    \scriptsize
    \begingroup
    \setlength{\tabcolsep}{2.2pt}
    \renewcommand{\arraystretch}{1.08}
    \renewcommand{\tabularxcolumn}[1]{m{#1}}
    \rowcolors{2}{gray!7}{white}
    \begin{tabularx}{\linewidth}{
        >{\centering\arraybackslash}m{0.040\linewidth}
        >{\raggedright\arraybackslash}m{0.235\linewidth}
        >{\centering\arraybackslash}m{0.052\linewidth}
        >{\centering\arraybackslash}m{0.064\linewidth}
        >{\centering\arraybackslash}m{0.064\linewidth}
        >{\raggedright\arraybackslash}m{0.170\linewidth}
        >{\raggedright\arraybackslash}X}
        \toprule
        \rowcolor{white}
        \textbf{ID} & \textbf{Task} & \textbf{Arm} & \textbf{Demos} & \textbf{Eval} & \textbf{Skill Type} & \textbf{Goal} \\
        \midrule
        T1 & Sequential Object Stacking & Single & 3,000 & 50 & Spatial alignment & Stack three baskets on the target mat. \\
        T2 & Articulated Container Manipulation & Single & 1,991 & 50 & Articulated manipulation & Insert a trash bag into a box and close the lid. \\
        T3 & Tool-Mediated Stamping & Single & 2,996 & 50 & Tool-mediated contact & Stamp a card using an ink pad. \\
        T4 & Precision Slot Insertion & Single & 2,001 & 50 & Fine-grained insertion & Insert remote controls into holder slots. \\
        T5 & Bimanual Material Pouring & Dual & 1,600 & 50 & Bimanual coordination & Pour beans into a receiving basket. \\
        T6 & Bimanual Packing and Transport & Dual & 1,600 & 50 & Cooperative transport & Pack objects and carry the basket to the target area. \\
        T7 & Dynamic Pick-and-Place & Dual & 2,012 & 50 & Dynamic grasping & Pick cans from a rotating turntable and place them on the target mats. \\
        T8 & Category Sorting and Placement & Dual & 1,600 & 50 & Semantic grounding & Sort four mahjong tiles into matching baskets. \\
        T9 & Long-Horizon Rearrangement & Dual & 1,600 & 50 & Long-horizon planning & Arrange cups, cans, and bread at target locations. \\
        T10 & Deformable Object Folding & Dual & 1,600 & 50 & Deformable manipulation & Fold pants into the target configuration. \\
        \bottomrule
    \end{tabularx}
    \endgroup
\end{table}

\noindent\textbf{Bimanual Task Coverage.}
The six bimanual tasks extend the benchmark beyond single-arm tabletop manipulation by introducing coupled hand roles and coordination requirements.
They require one arm to stabilize an object while the other performs a complementary action, timed interaction with moving objects, semantic sorting over multiple targets, deformable-object handling, and long-horizon sequencing across heterogeneous objects and receptacles.
These tasks therefore test failure modes that are difficult to expose with single-arm pick-place, tool-use, or insertion tasks alone.

\noindent\textbf{Seen and Unseen Conditions.}
Each task is evaluated under seen and unseen condition cells.
The seen condition uses factor values covered by the training demonstrations.
Factor A introduces held-out object instances, appearances, categories, or combinations.
Factor B introduces held-out positions, layouts, poses, or task dynamics.
Appearance variants include color, material, size, and texture changes.
Layout and pose variants include larger position shifts, denser object arrangements, target-region changes, distractors, new initial orientations, tilted poses, or changes in the orientation of functional parts.
This compact two-factor taxonomy makes the benchmark easier to interpret while preserving the main distribution shifts that matter for tabletop UMI manipulation.

\begin{table}[t]
    \centering
    \caption{Task-wise unseen evaluation shifts in UMI-Bench 1.0.
    Each task varies one object/category condition and one layout/motion condition during unseen evaluation.}
    \label{tab:task_generalization}
    \footnotesize
    \begingroup
    \setlength{\tabcolsep}{4pt}
    \renewcommand{\arraystretch}{1.05}
    \renewcommand{\tabularxcolumn}[1]{m{#1}}
    \begin{tabularx}{\linewidth}{
        >{\centering\arraybackslash}m{0.09\linewidth}
        >{\raggedright\arraybackslash}m{0.43\linewidth}
        >{\raggedright\arraybackslash}X
    }
        \toprule
        \textbf{\mbox{Task~ID}} 
        & \textbf{\mbox{Object / Category Shift}} 
        & \textbf{\mbox{Layout / Motion Shift}} \\
        \midrule
        T1  & Basket appearance & Stacking position \\
        T2  & Bag appearance & Object position \\
        T3  & Stamp, ink, and paper appearance & Stamping position \\
        T4  & Remote-control and holder appearance & Insertion position \\
        T5  & Basket--bean appearance & Basket--cup position \\
        T6  & Packing-object appearance & Box position \\
        T7  & Can and target-mat appearance & Turntable speed \\
        T8  & Mahjong category combination & Placement position \\
        T9  & Tabletop-object appearance & Placement position \\
        T10 & Pants appearance & Initial position \\
        \bottomrule
    \end{tabularx}
    \endgroup
\end{table}

\subsection{Evaluation Protocol and Metrics}
\label{sec:evaluation}

\noindent\textbf{Evaluation Episodes.}
Each task follows the 50-rollout real-world evaluation budget listed in Table~\ref{tab:task_suite}.
Rather than using a single seen/unseen split, each task defines two evaluation factors.
Factor A captures the primary object-, appearance-, category-, or combination-level shift, and Factor B captures the primary position-, layout-, pose-, or dynamics-level shift.
The cross product of these factors yields four cells: Seen/Seen, Seen/Unseen, Unseen/Seen, and Unseen/Unseen.
Task-specific factor definitions are listed in Table~\ref{tab:task_generalization}, and the resulting episode counts are summarized in Table~\ref{tab:rollout_distribution}.
All methods are evaluated on the same episode list.
The robot is reset according to the scene metadata before each episode.
The policy then runs until the task terminates or reaches the maximum rollout budget.

\begin{table}[t]
    \centering
    \caption{Evaluation rollout distribution for UMI-Bench 1.0.
    Cell labels follow the Factor-A/Factor-B order, where Seen denotes training-covered values and Unseen denotes held-out values.}
    \label{tab:rollout_distribution}
    \small
    \setlength{\tabcolsep}{4pt}
    \renewcommand{\arraystretch}{0.8}
    \begin{adjustbox}{max width=\linewidth}
    \begin{tabularx}{\linewidth}{
        >{\raggedright\arraybackslash}X
        >{\centering\arraybackslash}m{0.25\linewidth}
        >{\centering\arraybackslash}m{0.25\linewidth}
        >{\centering\arraybackslash}m{0.07\linewidth}
    }
        \toprule
        \textbf{Task name} 
        & {\footnotesize\textbf{Seen Objects} \textit{(Pos:S/U)}} 
        & {\footnotesize\textbf{Unseen Objects} \textit{(Pos:S/U)}} 
        & \textbf{Total} \\
        \midrule
        Sequential Object Stacking & 9 / 21 & 6 / 14 & 50 \\
        Articulated Container Manipulation & 10 / 20 & 6 / 14 & 50 \\
        Tool-Mediated Stamping & 20 / 10 & 14 / 6 & 50 \\
        Precision Slot Insertion & 14 / 16 & 10 / 10 & 50 \\
        Bimanual Material Pouring & 9 / 16 & 9 / 16 & 50 \\
        Bimanual Packing and Transport & 9 / 16 & 9 / 16 & 50 \\
        Dynamic Pick-and-Place & 20 / 10 & 10 / 10 & 50 \\
        Category Sorting and Placement & 9 / 16 & 9 / 16 & 50 \\
        Long-Horizon Rearrangement & 17 / 8 & 17 / 8 & 50 \\
        Deformable Object Folding & 13 / 12 & 13 / 12 & 50 \\
        \bottomrule
    \end{tabularx}
    \end{adjustbox}
\end{table}

\noindent\textbf{Episode Specification.}
Each real-robot evaluation trial is defined as one episode.
An episode specifies the initial state, task goal, execution budget, sensing and action interfaces, logging protocol, and outcome-assessment rule.
The initial state includes the robot start pose, object placement, and task-specific scene configuration.
For position generalization, objects are moved across predefined reset regions, producing seen and unseen placement conditions.
Task goals are specified by the task descriptions in Appendix~\ref{app:task_cards} and the detailed rubrics in Appendix~\ref{app:scoring_rubrics}.
An episode terminates when the task succeeds, the maximum step budget is reached, the scene clearly leaves the valid task range, or a safety-stop condition is triggered.
The outcome is then judged according to the task-specific rule and scored as success, failure, or partial completion.
The final benchmark results aggregate episode-level success rates and progress scores.
Implementation details for step budgets, action chunks, sensing rates, and rollout artifacts are provided in Appendix~\ref{app:evaluation_execution_details}.

\noindent\textbf{Evaluation Runner.}
The evaluation runner provides a unified interface between checkpoints and the robot workstation.
It loads a checkpoint together with the task and episode metadata, executes the shared action-space configuration, and records the rollout artifacts needed for audit.
Human score annotations are stored separately after rollout review, following the checklist in Appendix~\ref{app:evaluation_checklist}; the resulting package is sufficient for later inspection and leaderboard submission.

\noindent\textbf{Metrics.}
UMI-Bench reports two primary task metrics.
Full Success Rate (FSR) is the fraction of episodes that satisfy the full task criterion.
Progress Score is a partial-credit score on a 0--100 scale.
Task-specific subgoals are used to define progress rubrics, but are not reported as a separate primary metric in the main benchmark.
The first release does not report time-to-completion, retry count, failure-event count, action length, or execution-step count as official metrics.

\noindent\textbf{Shared Scoring Convention.}
All tasks are scored from the final stable physical state at rollout termination.
If a policy briefly fails a grasp or contact attempt but returns to the observation pose and completes the task within the budget, credit is assigned according to the final completed subgoals.
Placement credit is not awarded when an object is placed in the wrong target, wrong layer, or unstable configuration, or when a later action causes the object to fall, tip, or leave the valid region.
Full success requires all task-critical subgoals to be satisfied simultaneously in the final state.

\noindent\textbf{Generalization Analysis.}
Seen and unseen performance are reported separately.
The generalization gap is defined as $\Delta_{\mathrm{gen}} = S_{\mathrm{seen}} - S_{\mathrm{unseen}}$, where $S$ denotes Progress Score in the reported condition-level analysis, computed separately for Factor A and Factor B.
This analysis makes it possible to distinguish failures caused by object, appearance, or combination shifts from failures caused by spatial, geometric, or dynamic-condition shifts.

\section{Experiments}
\label{sec:experiments}


This section specifies the empirical protocol used in the real-robot evaluation.
The evaluated methods are $\pi_0$~\cite{pi_0}, $\pi_{0.5}$~\cite{pi_0_5}, and DreamZero~\cite{dreamzero}.
Unless otherwise specified, all methods share the same observation interface, action space, task definitions, training data splits, and evaluation episode lists.


The experiments are organized around four questions.
First, we ask how current policies perform on the overall UMI-Bench task suite.
Second, we examine which tasks reveal model-specific strengths and failure modes.
Third, we measure how performance changes across Seen/Seen, Factor-A, Factor-B, and combined-shift evaluation conditions.
Fourth, we analyze what a standardized UMI-style data and evaluation protocol makes visible beyond aggregate success rates.

\subsection{Main Results}

Table~\ref{tab:main_results} reports condition-level real-robot scores for the four rollout cells defined in Tables~\ref{tab:rollout_distribution} and~\ref{tab:detailed_split_definitions}.
Averaged over the ten tasks, $\pi_{0.5}$ achieves the best Overall Score, with 55.84 compared to 48.90 for $\pi_0$ and 40.59 for DreamZero.
Its advantage is broad rather than task-specific, ranking first on six of the ten tasks and covering both single-arm and bimanual settings.
In contrast, $\pi_0$ performs best on the two single-arm placement tasks T1 and T2, while DreamZero is competitive on T3 and T6 but trails on average.
The hardest tasks are T3 and T9, where all methods obtain 0\% FSR, indicating that long-horizon stage estimation and tight final-placement accuracy remain challenging.

Across all tasks and models, performance drops more under Factor-B shifts than under Factor-A shifts.
The mean Seen/Seen score is 59.62, decreasing to 53.45 under Factor-A shifts, 45.33 under Factor-B shifts, and 40.19 under combined shifts.
This suggests that current policies retain much of their capability when training and evaluation distributions are aligned, but still rely substantially on demonstration-aligned motion priors when task-relevant geometry or dynamics change.
Additional task-level diagnostics are provided in Appendix~\ref{app:additional_result_diagnostics}.

\begin{table}[!t]
    \centering
    \caption{\textbf{Primary task-wise result summary for UMI-Bench 1.0.}
    Score columns report the mean 0--100 Progress Score under the four evaluation cells defined in Table~\ref{tab:rollout_distribution}, and FSR denotes Full Success Rate.
    The Overall Score and Overall FSR columns aggregate all 50 real-world rollouts for each task--model pair.
    Each cell lists the three models in the order $\pi_0$\,/\,$\pi_{0.5}$\,/\,DZ, where DZ denotes DreamZero. \textbf{Bold} marks the best model per task per column.
    Column headers use object-position terminology as shorthand; tasks whose axes differ follow the task-specific Factor-A and Factor-B definitions in Table~\ref{tab:rollout_distribution}.}
    \label{tab:main_results}

    \scriptsize
    \setlength{\tabcolsep}{1.4pt}
    \renewcommand{\arraystretch}{1.15}
    \newcommand{\ressep}{\,/\,}
    \newcommand{\blocksep}{\vrule width 0.25pt}

    \resizebox{\linewidth}{!}{%
    \begin{tabular}{@{}lc!{\blocksep}c!{\blocksep}c!{\blocksep}c!{\blocksep}c!{\blocksep}c@{}}
        \toprule
        \textbf{Task ID} 
        & \textbf{Seen Obj.} 
        & \textbf{Seen Obj.} 
        & \textbf{Unseen Obj.} 
        & \textbf{Unseen Obj.} 
        & \textbf{Overall} 
        & \textbf{Overall} \\
        & \textbf{Seen Pos.} 
        & \textbf{Unseen Pos.} 
        & \textbf{Seen Pos.} 
        & \textbf{Unseen Pos.} 
        & \textbf{Score} 
        & \textbf{FSR (\%)} \\
        \cmidrule(lr){2-7}
        & $\pi_0$\ressep$\pi_{0.5}$\ressep DZ
        & $\pi_0$\ressep$\pi_{0.5}$\ressep DZ
        & $\pi_0$\ressep$\pi_{0.5}$\ressep DZ
        & $\pi_0$\ressep$\pi_{0.5}$\ressep DZ
        & $\pi_0$\ressep$\pi_{0.5}$\ressep DZ
        & $\pi_0$\ressep$\pi_{0.5}$\ressep DZ \\
        \midrule

        T1 & 85.00\ressep\textbf{86.11}\ressep72.22 & \textbf{70.48}\ressep68.33\ressep57.14 & \textbf{100.00}\ressep83.33\ressep73.33 & \textbf{64.64}\ressep59.64\ressep58.21 & \textbf{75.00}\ressep70.90\ressep62.10 & \textbf{34.00\%}\ressep14.00\%\ressep2.00\% \\
        T2 & 75.50\ressep\textbf{82.50}\ressep73.50 & \textbf{87.50}\ressep79.75\ressep80.75 & \textbf{95.00}\ressep92.50\ressep54.17 & \textbf{89.29}\ressep69.29\ressep48.21 & \textbf{86.50}\ressep78.90\ressep67.00 & \textbf{58.00\%}\ressep\textbf{58.00\%}\ressep22.00\% \\
        T3 & 25.50\ressep22.50\ressep\textbf{51.50} & 21.00\ressep24.50\ressep\textbf{34.00} & 22.14\ressep\textbf{25.71}\ressep17.50 & \textbf{21.67}\ressep14.17\ressep16.67 & 23.20\ressep22.80\ressep\textbf{34.30} & \textbf{0.00\%}\ressep\textbf{0.00\%}\ressep\textbf{0.00\%} \\
        T4 & 35.71\ressep\textbf{61.43}\ressep51.43 & 35.63\ressep\textbf{50.00}\ressep38.13 & 39.00\ressep\textbf{60.00}\ressep57.00 & 32.00\ressep31.00\ressep\textbf{47.00} & 35.60\ressep\textbf{51.40}\ressep47.40 & 6.00\%\ressep\textbf{20.00\%}\ressep0.00\% \\
        T5 & \textbf{86.67}\ressep85.56\ressep20.00 & 22.81\ressep\textbf{35.00}\ressep20.00 & 73.89\ressep\textbf{74.44}\ressep24.44 & 10.94\ressep\textbf{34.06}\ressep24.69 & 39.70\ressep\textbf{50.90}\ressep22.30 & 22.00\%\ressep\textbf{26.00\%}\ressep0.00\% \\
        T6 & 78.89\ressep88.89\ressep\textbf{97.78} & 22.50\ressep\textbf{71.56}\ressep66.88 & 51.67\ressep78.89\ressep\textbf{94.44} & 38.75\ressep\textbf{51.56}\ressep\textbf{51.56} & 43.10\ressep69.60\ressep\textbf{72.50} & 22.00\%\ressep40.00\%\ressep\textbf{52.00\%} \\
        T7 & 61.00\ressep\textbf{80.00}\ressep13.50 & 69.00\ressep\textbf{84.00}\ressep31.00 & \textbf{33.00}\ressep24.50\ressep7.50 & 74.00\ressep\textbf{88.00}\ressep10.50 & 59.60\ressep\textbf{71.30}\ressep15.20 & 18.00\%\ressep\textbf{38.00\%}\ressep0.00\% \\
        T8 & \textbf{86.11}\ressep75.00\ressep27.78 & 31.56\ressep\textbf{38.75}\ressep36.25 & 80.56\ressep\textbf{86.11}\ressep33.89 & 38.75\ressep\textbf{42.81}\ressep30.94 & 52.50\ressep\textbf{55.10}\ressep32.60 & \textbf{22.00\%}\ressep18.00\%\ressep0.00\% \\
        T9 & 25.29\ressep\textbf{31.76}\ressep9.41 & 6.25\ressep\textbf{18.75}\ressep8.75 & 26.47\ressep\textbf{32.35}\ressep11.76 & 6.25\ressep10.00\ressep\textbf{12.50} & 19.60\ressep\textbf{26.40}\ressep10.60 & \textbf{0.00\%}\ressep\textbf{0.00\%}\ressep\textbf{0.00\%} \\
        T10 & \textbf{74.62}\ressep67.69\ressep55.77 & 52.50\ressep\textbf{55.83}\ressep41.25 & 47.31\ressep\textbf{62.69}\ressep39.23 & 41.25\ressep\textbf{57.08}\ressep30.42 & 54.20\ressep\textbf{61.00}\ressep41.90 & 8.00\%\ressep\textbf{22.00\%}\ressep0.00\% \\

        \bottomrule
    \end{tabular}%
    }
\end{table}

\begin{figure}[t]
    \centering
    \captionsetup[subfigure]{skip=2pt}
    \captionsetup{skip=3pt}

    \begingroup
    \normalsize
    \textcolor[HTML]{3F5E7D}{\rule{0.9em}{0.65em}}\hspace{0.3em}$\bm{\pi_0}$
    \hspace{1.0em}
    \textcolor[HTML]{D7B46A}{\rule{0.9em}{0.65em}}\hspace{0.3em}$\bm{\pi_{0.5}}$
    \hspace{1.0em}
    \textcolor[HTML]{BF6B73}{\rule{0.9em}{0.65em}}\hspace{0.3em}\textbf{DreamZero}
    \par\vspace{-0.2em}
    \endgroup

    \begin{subfigure}[t]{0.485\linewidth}
        \centering
        \includegraphics[width=\linewidth]{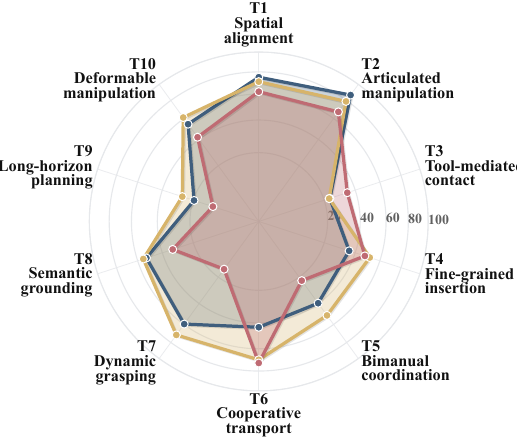}
        \caption{Capability-wise Overall Score.}
        \label{fig:taskwise_progress}
    \end{subfigure}
    \hspace{0.015\linewidth}
    \begin{subfigure}[t]{0.485\linewidth}
        \centering
        \includegraphics[width=\linewidth]{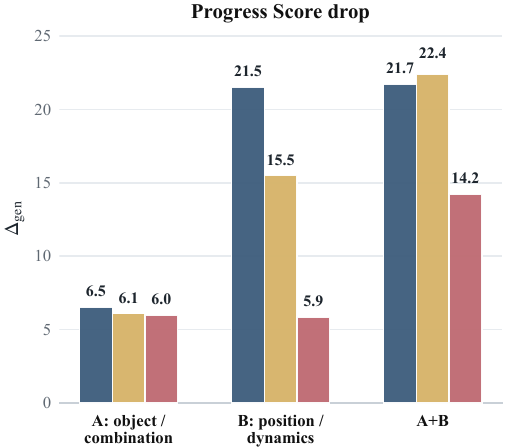}
        \caption{Generalization Gap in Progress Score.}
        \label{fig:generalization}
    \end{subfigure}

    \caption{\textbf{Overall performance and generalization summary for UMI-Bench 1.0.}
  Left: Overall Score grouped by skill type.
  Right: generalization gap in Progress Score relative to the Seen/Seen condition. Factor A denotes object-, appearance-, category-, or combination-level shifts, whereas Factor B denotes position-, layout-, pose-, or dynamics-level shifts.}
    \label{fig:result_summary_plots}
\end{figure}

\subsection{Protocol Implications}

The result patterns above also clarify the intended role of UMI-Bench in the benchmark design space.
UMI-Bench does not aim to replace remote robot platforms, which primarily improve public access to shared machines.
Instead, it emphasizes reproducible local evaluation for a specific data and deployment paradigm.
This makes the benchmark suitable for iterative model development, where teams repeatedly train policies, test them under controlled real-world perturbations, inspect failures, and update their data collection strategy.

UMI-Bench also differs from pure dataset releases.
It treats data collection, scene specification, evaluation episodes, and scoring as parts of one unified protocol.
This alignment is important for UMI-style policies because their performance depends on the coupling between wrist-view observations, action chunks, and physical scene reconstruction.
By recording scene metadata and result packages, UMI-Bench makes this coupling explicit and analyzable.

\section{Conclusion}
\label{sec:conclusion}

We presented UMI-Bench 1.0, a local-first real-robot benchmark for UMI-style wrist-view manipulation policies.
UMI-Bench standardizes data collection, hardware setup, scene reset, evaluation execution, result packaging, and task-factor analysis.
The first release contains 10 tabletop tasks, approximately 20k demonstrations, and 50 evaluation episodes per task.
By aligning UMI data collection with reproducible real-robot evaluation, UMI-Bench provides a practical benchmark infrastructure for measuring and diagnosing real-world manipulation policies.
To support reproducible local evaluation of UMI-style policies, we will release the benchmark protocol, episode specifications, scene metadata, scoring rubrics, evaluation runner, baseline checkpoints/logs, and rollout audit packages.

\section{Limitation}
\label{sec:Limitation}

UMI-Bench 1.0 has several limitations. First, the current benchmark focuses on tabletop manipulation, so it does not yet cover mobile manipulation, large workspaces, or more complex long-horizon scenarios. Second, although the first release includes approximately 20k demonstrations, the data scale remains limited compared with large robot datasets. Third, real-world evaluation can still be affected by operator reset variation, calibration drift, lighting changes, and hardware timing effects. 
\FloatBarrier
\bibliographystyle{plainnat}
\bibliography{ref}

\clearpage
\appendix

\section{Standardized Hardware Platform}
\label{app:hardware}

This appendix records the workstation configuration needed to reproduce the UMI-Bench 1.0 evaluation setup.

\begin{table}[H]
    \centering
    \caption{Standardized workstation specification used in UMI-Bench 1.0.}
    \label{tab:hardware_spec}
    \footnotesize
    \begingroup
    \setlength{\tabcolsep}{4pt}
    \renewcommand{\arraystretch}{1.08}
    \begin{tabularx}{\linewidth}{>{\raggedright\arraybackslash}m{0.24\linewidth}>{\raggedright\arraybackslash}X}
        \toprule
        \textbf{Component} & \textbf{Configuration} \\
        \midrule
        Manipulator & FastTouch tabletop arm with a released 6-DoF URDF and Python SDK. The SDK exposes joint, Cartesian pose, waypoint, IK, and gripper-control APIs. \\
        Control loop & Waypoint trajectories are internally sampled at 400 Hz. Gripper command and state handling use a 5 ms period. \\
        End effector & Robot-side gripper controlled through the FastTouch SDK gripper-control interface. Evaluation logs record commanded gripper values and observed gripper state together with the robot pose. \\
        Evaluation wrist-view sensing & Wrist RGB cameras provide 1280 by 1280 images at 100 FPS. Policy inputs are resized to 224 by 224, and robot state is logged at 30 Hz by default. \\
        Localization and reset & The physical workstation uses a fixed 1.2 m by 1.0 m by 0.75 m tabletop. Objects are reset on an acrylic grid board matching the 1.2 m by 1.0 m tabletop footprint; the board is divided into 5 cm by 5 cm cells with operator-visible grid labels for repeatable placement and metadata recording. \\
        Runtime interface & FastTouch evaluation uses the FastTouch SDK runtime libraries for Ubuntu 20.04/22.04/24.04. \\
        \bottomrule
    \end{tabularx}
    \endgroup
\end{table}

The reset surface is a transparent acrylic board placed on top of the physical tabletop.
The board footprint matches the 1.2 m by 1.0 m tabletop, and the printed grid uses 5 cm by 5 cm cells.
During collection and evaluation, objects are placed on this acrylic grid rather than directly on the table surface.
The grid coordinates provide a repeatable reset reference for operators and for scene metadata, but they are not treated as policy-readable visual markers.

Detailed joint limits, URDF-level gripper kinematics, topic-rate checks, and host-computer settings are archived with the supplementary workspace notes.

\section{FastUMI Pro Collection Workflow}
\label{app:fastumi_pro}

FastUMI Pro is used as the collection-side sensing and logging toolchain.
It records demonstration-side RGB, TOF, IMU, gripper, and spatial-trajectory streams.
The workflow below summarizes the checks performed before a session is accepted into the benchmark repository.

\begin{table}[H]
    \centering
    \caption{FastUMI Pro collection workflow.}
    \label{tab:fastumi_workflow}
    \footnotesize
    \begingroup
    \setlength{\tabcolsep}{4pt}
    \renewcommand{\arraystretch}{1.08}
    \begin{tabularx}{\linewidth}{>{\raggedright\arraybackslash}m{0.22\linewidth}>{\raggedright\arraybackslash}X}
        \toprule
        \textbf{Stage} & \textbf{Protocol summary} \\
        \midrule
        Initialization & Configure the FastUMI SDK and ROS environment, connect the device, and launch the device driver. Multi-device sessions apply the USB bandwidth setup before recording. \\
        Stream check & Verify pose, RGB, TOF, IMU, and gripper streams with the monitor tool and RViz. Reference checks include 500 Hz pose or IMU sampling, 60 Hz RGB, 30 Hz TOF, and dynamic gripper readings. \\
        Calibration and reset & Confirm camera and robot-frame configuration, select the task and episode metadata, and reconstruct the scene from the reset image and tabletop markers. \\
        Recording & Save wrist-view video, timestamps, gripper state, spatial trajectory, scene metadata, failure notes, and quality flags during the operator demonstration. \\
        Quality control & Check stream completeness, video readability, timestamp consistency, and task-rule compliance. Invalid or ambiguous sessions are marked before export. \\
        Export & Convert accepted sessions into HDF5 or LeRobot-compatible files, with automatic single- or dual-arm layout detection and alignment to the target training frequency. \\
        \bottomrule
    \end{tabularx}
    \endgroup
\end{table}

\section{Data Schema}
\label{app:data_schema}

Each episode contains episode metadata, scene JSON, action chunk records, robot state records, camera frame references, quality flags, and score JSON.
The scene JSON contains task identifier, object identifier, object category, appearance, material, tabletop marker position, pose, target region, split, and reset image.
The action schema specifies coordinate frame, units, chunk length, gripper encoding, timestamps, and execution status.
Raw FastUMI sessions store RGB video and timestamps, clamp data, and merged trajectory logs.
Dual-arm sessions use separate left-hand and right-hand folders with the same per-arm structure.
During conversion, 60 fps raw streams are aligned to the requested 20, 30, or 60 Hz training frequency.
The HDF5 export stores images as uint8 tensors and uses eight-dimensional state/action vectors of the form $[x,y,z,q_x,q_y,q_z,q_w,\mathrm{clamp}]$ for each arm stream.

\subsection{Evaluation Execution Details}
\label{app:evaluation_execution_details}

Each evaluation episode uses a fixed maximum step budget, set from the nominal time required to complete a standard execution with additional tolerance for minor retries such as a missed grasp or intermediate correction.
The maximum step budgets for T1--T10 are 1000, 1200, 1000, 800, 1200, 900, 1000, 1000, 1800, and 1000 steps, respectively.
Policies are executed through action chunks rather than individual low-level commands: each inference call predicts a waypoint trajectory, which is then executed by the robot SDK.
Single-arm $\pi_0$, single-arm DreamZero, and dual-arm DreamZero use up to 24 waypoints by default.
The dual-arm $\pi_0$ test runner uses up to 50 waypoints by default, and uses 20 waypoints for long-horizon and dynamic tasks.
The observation input is provided by wrist cameras at 1280$\times$1280 resolution and 100 FPS, with policy images resized to 224$\times$224.
Robot states are logged at 30 Hz by default.

The evaluation runner takes a checkpoint, task configuration, episode list, scene JSON, calibration files, and action-space configuration as input.
During each rollout, it records wrist-camera observations, robot states, action chunks, timestamps, and runtime logs.
After each rollout, the system stores a manifest, action records, state samples, a trajectory plot, and optional video.
The manifest records the task, model, hardware, timing, execution status, step count, runtime frequencies, and output paths; action records store predicted targets, transformed target poses, gripper commands, timestamps, and step indices; and state samples record the actual robot motion for later inspection.
Human score annotations are stored separately after rollout review.

\paragraph{Inference latency.}
\label{app:inference_latency}
We also profile the average inference latency of each policy under its evaluation hardware.
On a single NVIDIA A100 GPU, $\pi_{0.5}$ and $\pi_0$ have inference latencies of 219.3 ms and 198.8 ms, respectively.
DreamZero has an inference latency of 1630.0 ms on two NVIDIA A100 GPUs.
These numbers are reported as deployment-efficiency references rather than hardware-normalized comparisons.

\section{Task Cards}
\label{app:task_cards}

\begin{figure}[tbp]
    \centering
    \includegraphics[max width=\linewidth]{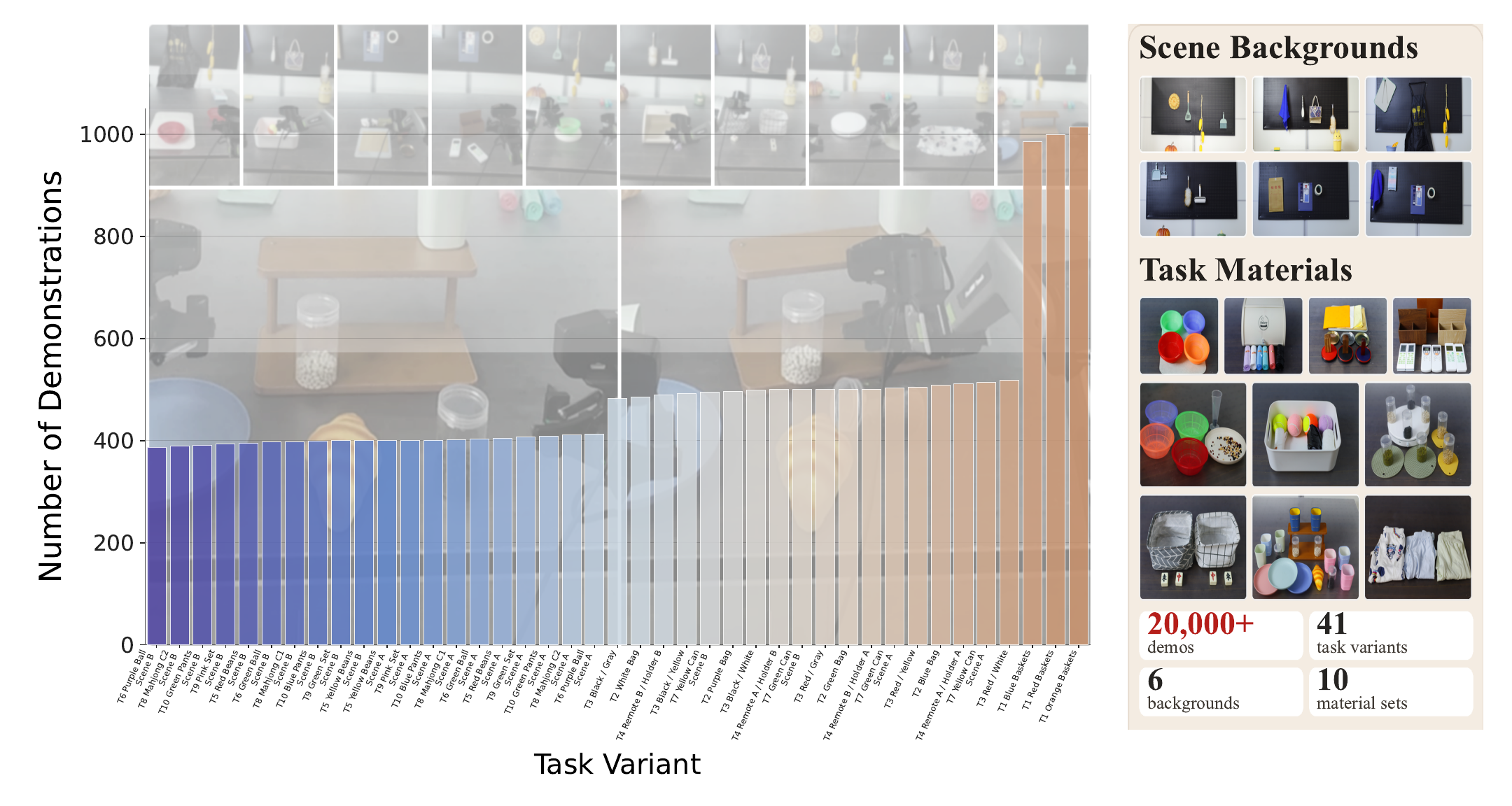}
    \caption{\textbf{Task and variant coverage in UMI-Bench 1.0.}
    The figure summarizes the distribution of benchmark task variants across the 10-task suite and shows representative real-world scene examples used in data collection and evaluation.}
    \label{fig:task_variant_distribution}
\end{figure}

Each task card contains the task prompt, object list, reset rule, scoring stages, seen episodes, and Factor-A/Factor-B held-out conditions.
Each task card also states whether the task stresses target selection, spatial reasoning, grasp stability, placement precision, contact interaction, or multi-stage execution.
This appendix records the task cards for all ten tasks used in the first release.
Figure~\ref{fig:task_variant_distribution} summarizes variant-level demonstration coverage together with representative scene backgrounds and task materials.
This visualization complements the task-level summary in Table~\ref{tab:task_suite}.
Table~\ref{tab:detailed_split_definitions} lists the concrete evaluation split definitions used to instantiate the rollout cells in Table~\ref{tab:rollout_distribution}.

\begin{table}[H]
    \centering
    \caption{Detailed evaluation split definitions for UMI-Bench 1.0.
    The first factor captures object, appearance, category-combination, or material variation; the second factor captures position, layout, or speed variation.}
    \label{tab:detailed_split_definitions}
    \scriptsize
    \setlength{\tabcolsep}{2.4pt}
    \renewcommand{\arraystretch}{1.12}
    \begin{tabularx}{\linewidth}{>{\centering\arraybackslash}p{0.075\linewidth}>{\raggedright\arraybackslash}X>{\raggedright\arraybackslash}X>{\raggedright\arraybackslash}X>{\raggedright\arraybackslash}X}
        \toprule
        \textbf{Task ID} & \textbf{Factor A Seen} & \textbf{Factor A Unseen} & \textbf{Factor B Seen} & \textbf{Factor B Unseen} \\
        \midrule
        T1
        & Three same-color baskets
        & Three mixed-color baskets
        & Position 1
        & Positions 2 and 3 \\
        T2
        & White, purple, blue, and green garbage bags
        & Red garbage bag and large black garbage bag
        & Position-angle 2
        & Position-angles 1 and 3 \\
        T3
        & Red or black stamp with white, gray, or yellow paper
        & Held-out stamp paste or held-out orange or brown paper
        & First two positions in each three-rollout cycle
        & Third position in each three-rollout cycle \\
        T4
        & Left remote, right remote, and holder types 1 or 2
        & Any type-3 left remote, right remote, or holder
        & Remote-holder layout position 1
        & Remote-holder layout position 2 \\
        T5
        & Orange basket with red or yellow beans
        & Red, blue, or green basket with black or white beans
        & Center $3\times3$ basket--measuring-cup layouts in the $5\times5$ grid
        & Outer 16 basket--measuring-cup layouts in the $5\times5$ grid \\
        T6
        & Green mesh ball, purple mesh ball, and tissue roll
        & Pink mesh ball, mixed-color mesh ball, white garbage bag, or black garbage bag
        & Center $3\times3$ cells of the $5\times5$ grid
        & Outer 16 cells of the $5\times5$ grid \\
        T7
        & Green and yellow cans
        & White and black cans
        & Medium turntable speed, 15 deg/s
        & Low speed, 8 deg/s, or fast speed, 30 deg/s \\
        T8
        & Seen mahjong color combination with crossed black/red arrangement
        & Unseen mahjong color combination with same-color left and right groups
        & Center $3\times3$ cells of the $5\times5$ grid
        & Outer 16 cells of the $5\times5$ grid \\
        T9
        & Dark-blue cup with green plate, or pink cup-plate set
        & Light-blue cup with dark-blue plate, or green special-shaped cup with dark-blue plate
        & Middle cross positions in the $3\times3$ placement region
        & Four corner positions in the $3\times3$ placement region \\
        T10
        & Green and blue pants
        & Patterned pants
        & Upper two rows of the $3\times3$ placement grid
        & Lower row of the $3\times3$ placement grid \\
        \bottomrule
    \end{tabularx}
\end{table}

For Precision Slot Insertion, the rollout sheet uses a four-digit condition code in the order of layout position, left-remote type, right-remote type, and holder type; the table above states this code in task-level terms.

\subsection{Single-Arm Rollout-Level Score Heatmaps}
\label{app:single_arm_heatmaps}

\begin{figure}[tbp]
    \centering
    \includegraphics[max width=\linewidth]{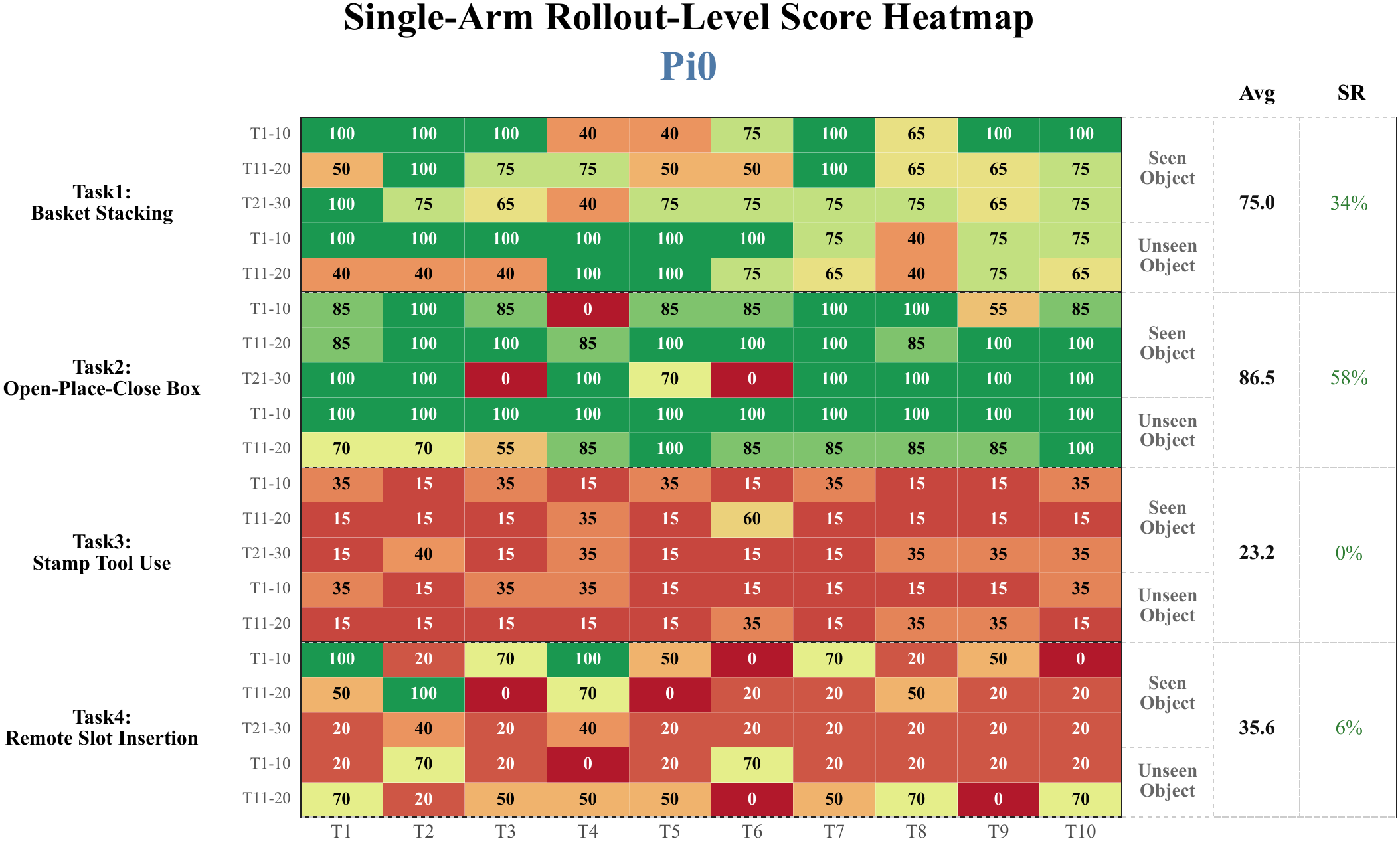}
    \caption{Single-arm rollout-level score heatmap for $\pi_0$ on Tasks 1--4.}
    \label{fig:single_arm_heatmap_pi0}
\end{figure}

\begin{figure}[tbp]
    \centering
    \includegraphics[max width=\linewidth]{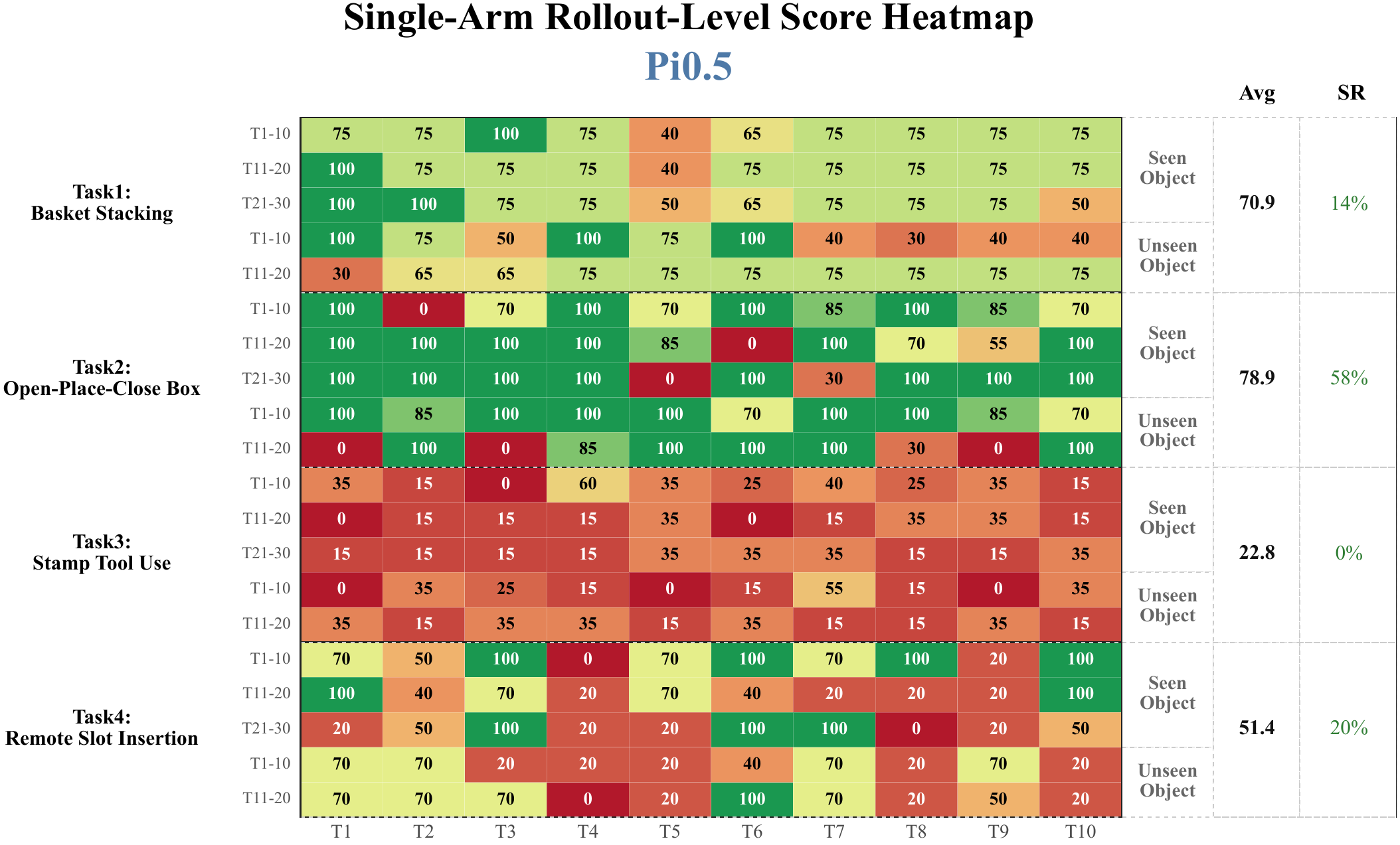}
    \caption{Single-arm rollout-level score heatmap for $\pi_{0.5}$ on Tasks 1--4.}
    \label{fig:single_arm_heatmap_pi05}
\end{figure}

\begin{figure}[tbp]
    \centering
    \includegraphics[max width=\linewidth]{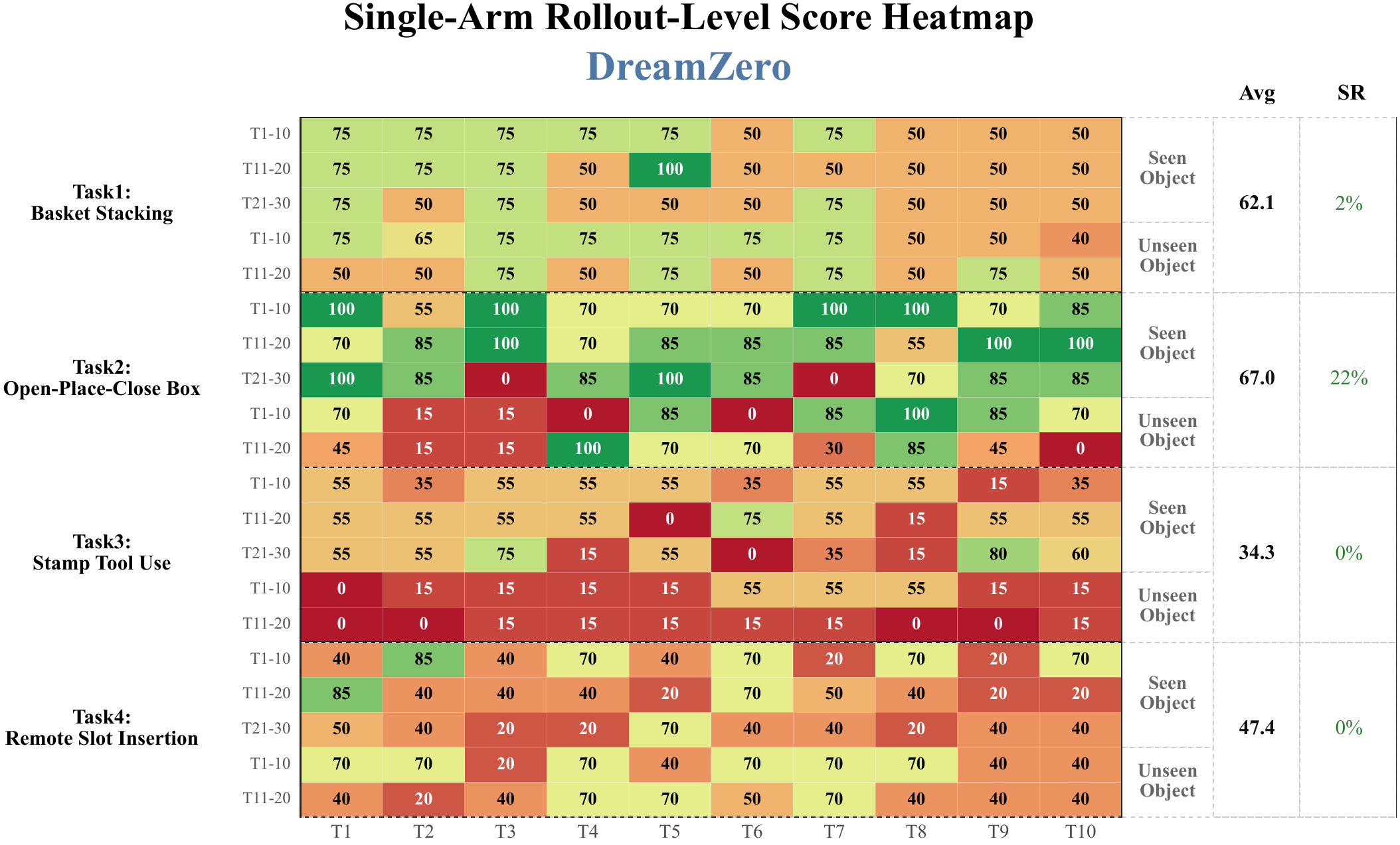}
    \caption{Single-arm rollout-level score heatmap for DreamZero on Tasks 1--4.}
    \label{fig:single_arm_heatmap_dreamzero}
\end{figure}

Figures~\ref{fig:single_arm_heatmap_pi0}--\ref{fig:single_arm_heatmap_dreamzero} show the per-rollout progress scores for the four single-arm tasks.
Rows group the evaluated object split, while columns preserve the rollout order and task-specific position or condition cells.

\subsection{Bimanual Rollout-Level Score Heatmaps}
\label{app:bimanual_heatmaps}

\begin{figure}[tbp]
    \centering
    \includegraphics[max width=\linewidth]{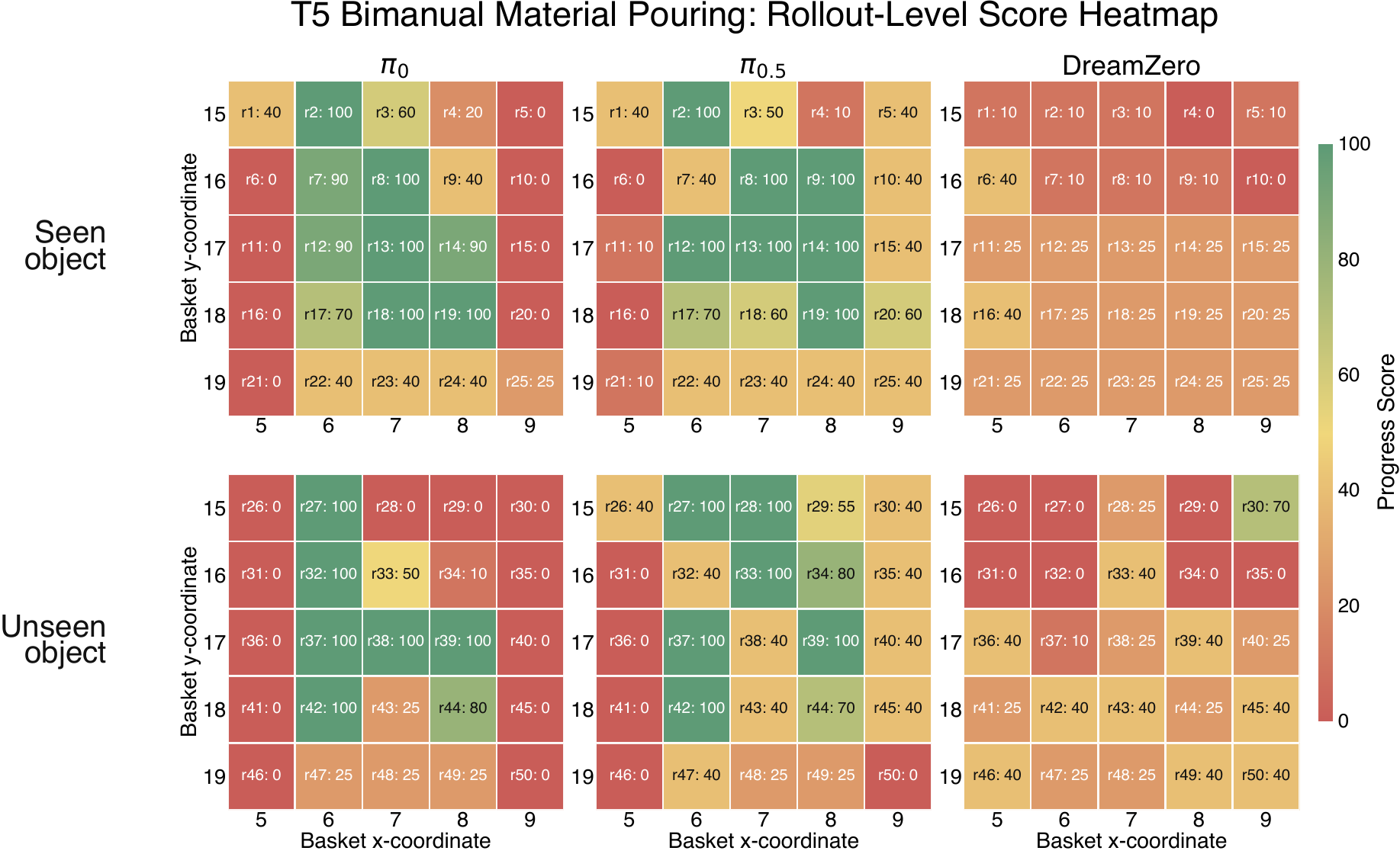}
    \caption{Task 5, Bimanual Material Pouring. Heatmap cells are physical basket--measuring-cup layouts; entries report rollout ID and Progress Score.}
    \label{fig:bimanual_heatmap_t5}
\end{figure}

\begin{figure}[tbp]
    \centering
    \includegraphics[max width=\linewidth]{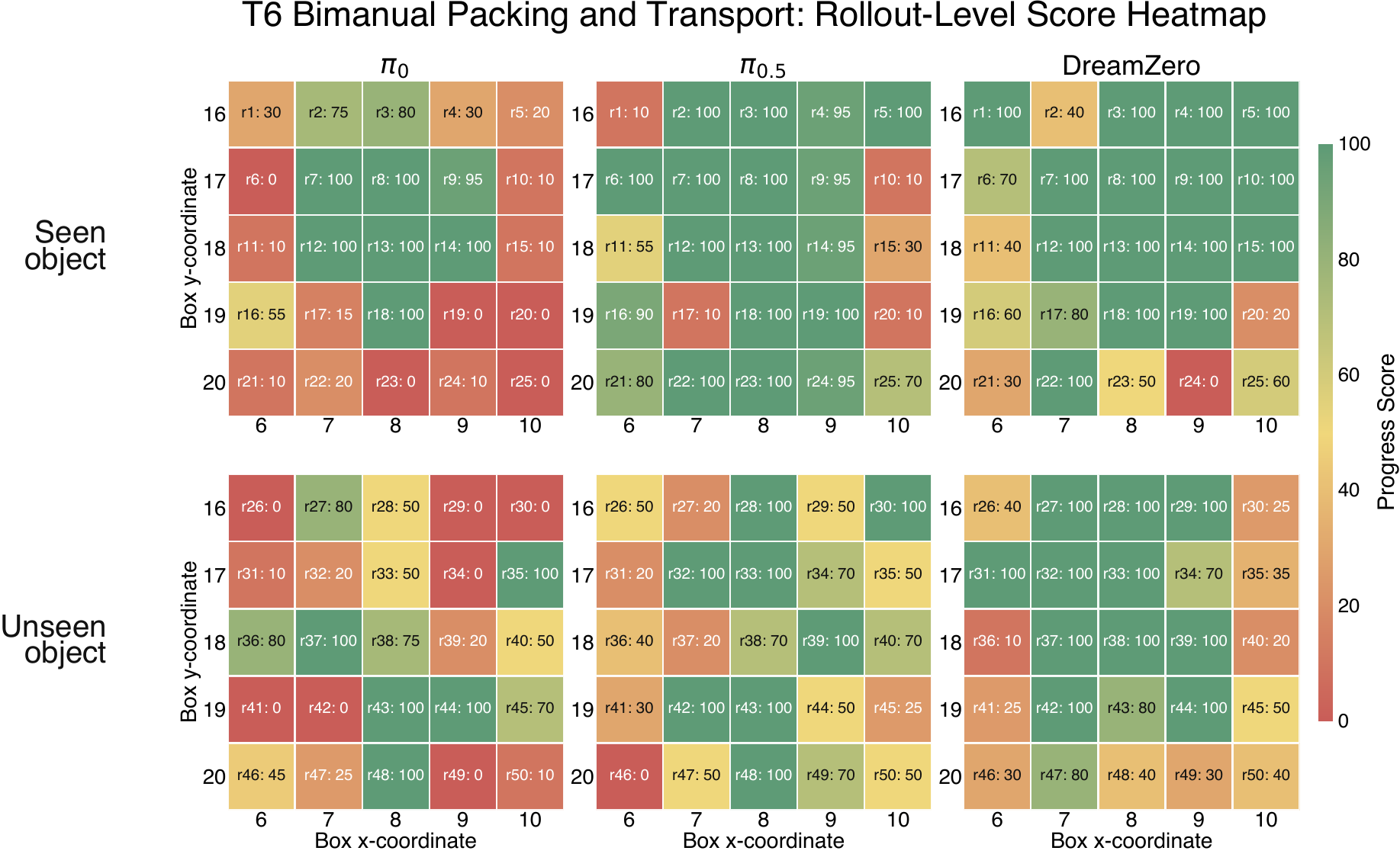}
    \caption{Task 6, Bimanual Packing and Transport. Heatmap cells are physical box coordinates; entries report rollout ID and Progress Score.}
    \label{fig:bimanual_heatmap_t6}
\end{figure}

\begin{figure}[tbp]
    \centering
    \includegraphics[max width=\linewidth]{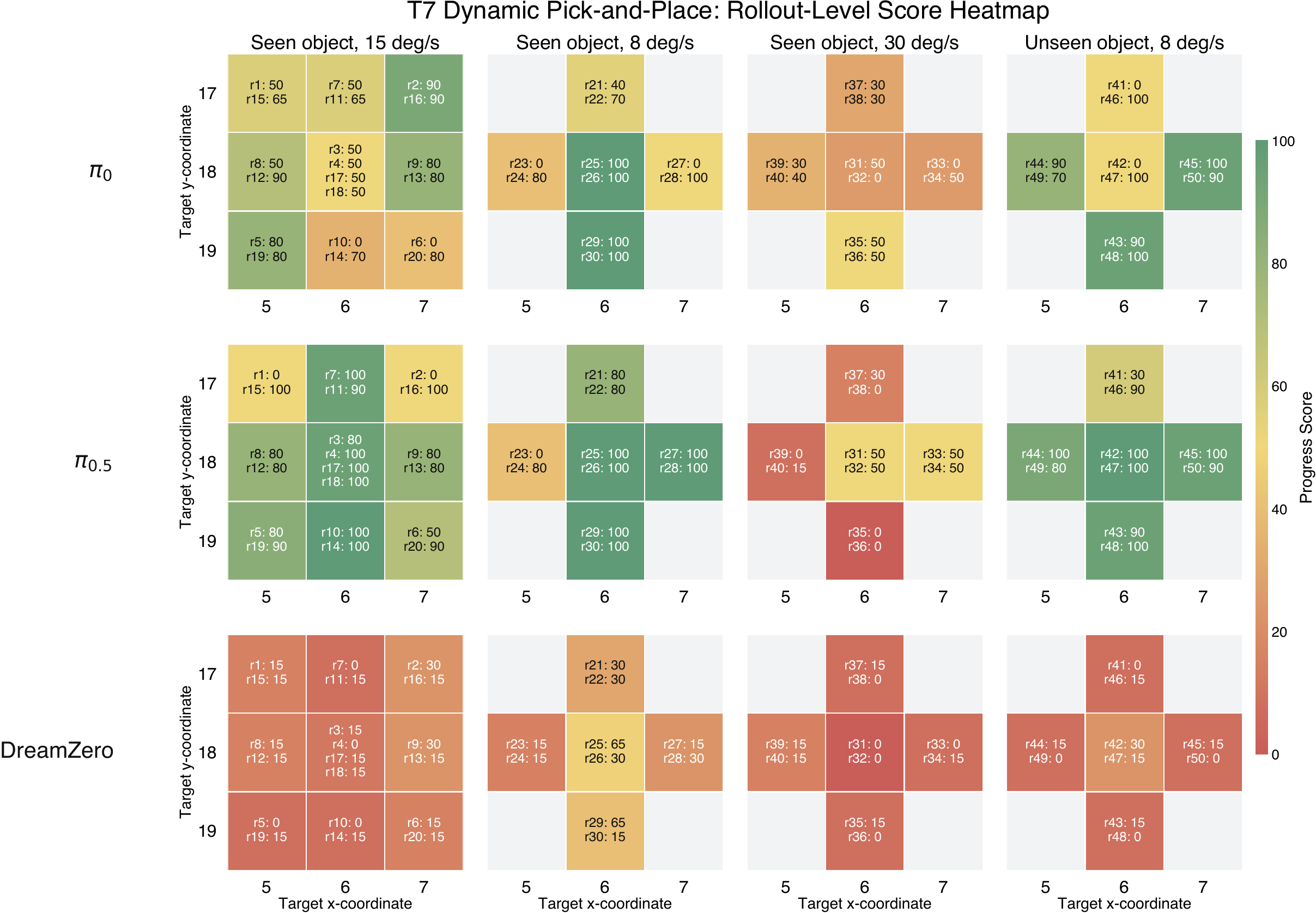}
    \caption{Task 7, Dynamic Pick-and-Place. Panels separate the evaluated turntable speed and object-appearance conditions; entries report rollout ID and Progress Score.}
    \label{fig:bimanual_heatmap_t7}
\end{figure}

\begin{figure}[tbp]
    \centering
    \includegraphics[max width=\linewidth]{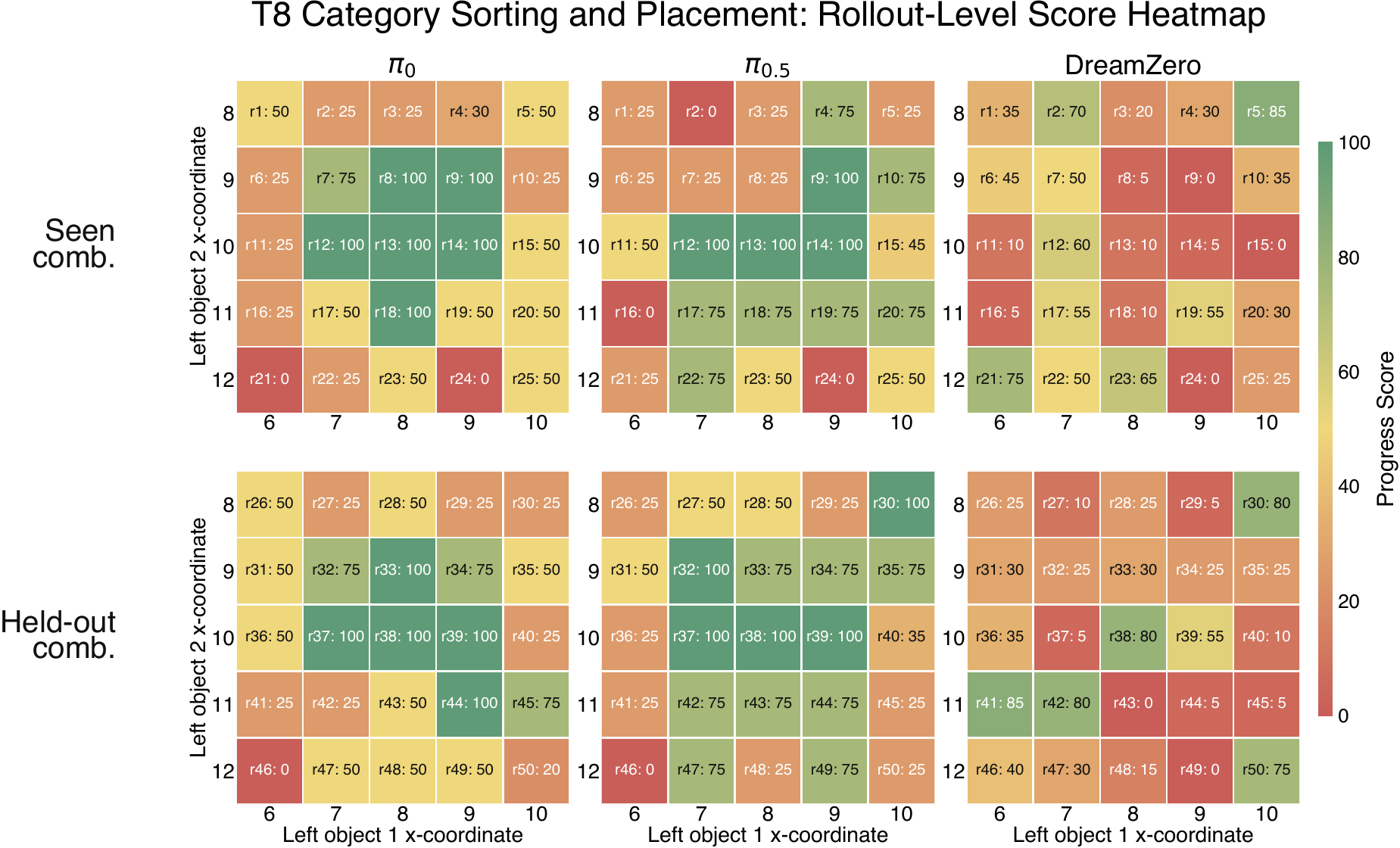}
    \caption{Task 8, Category Sorting and Placement. Heatmap cells follow the relative mahjong-placement grid; panels separate seen and held-out category combinations, and entries report rollout ID and Progress Score.}
    \label{fig:bimanual_heatmap_t8}
\end{figure}

\begin{figure}[tbp]
    \centering
    \includegraphics[max width=\linewidth]{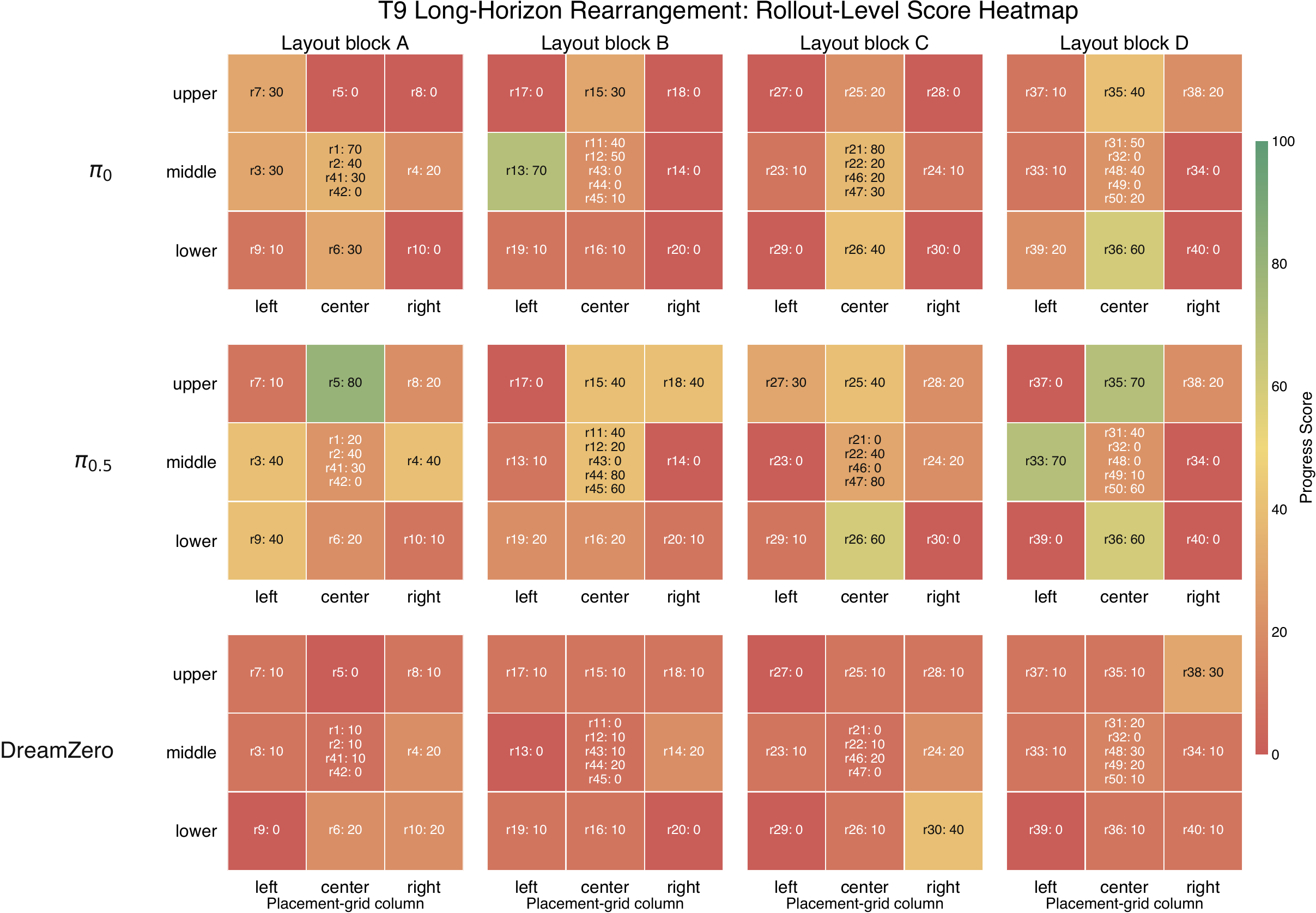}
    \caption{Task 9, Long-Horizon Rearrangement. Panels show the task-specific placement-grid condition blocks; entries report rollout ID and Progress Score.}
    \label{fig:bimanual_heatmap_t9}
\end{figure}

\begin{figure}[tbp]
    \centering
    \includegraphics[max width=\linewidth]{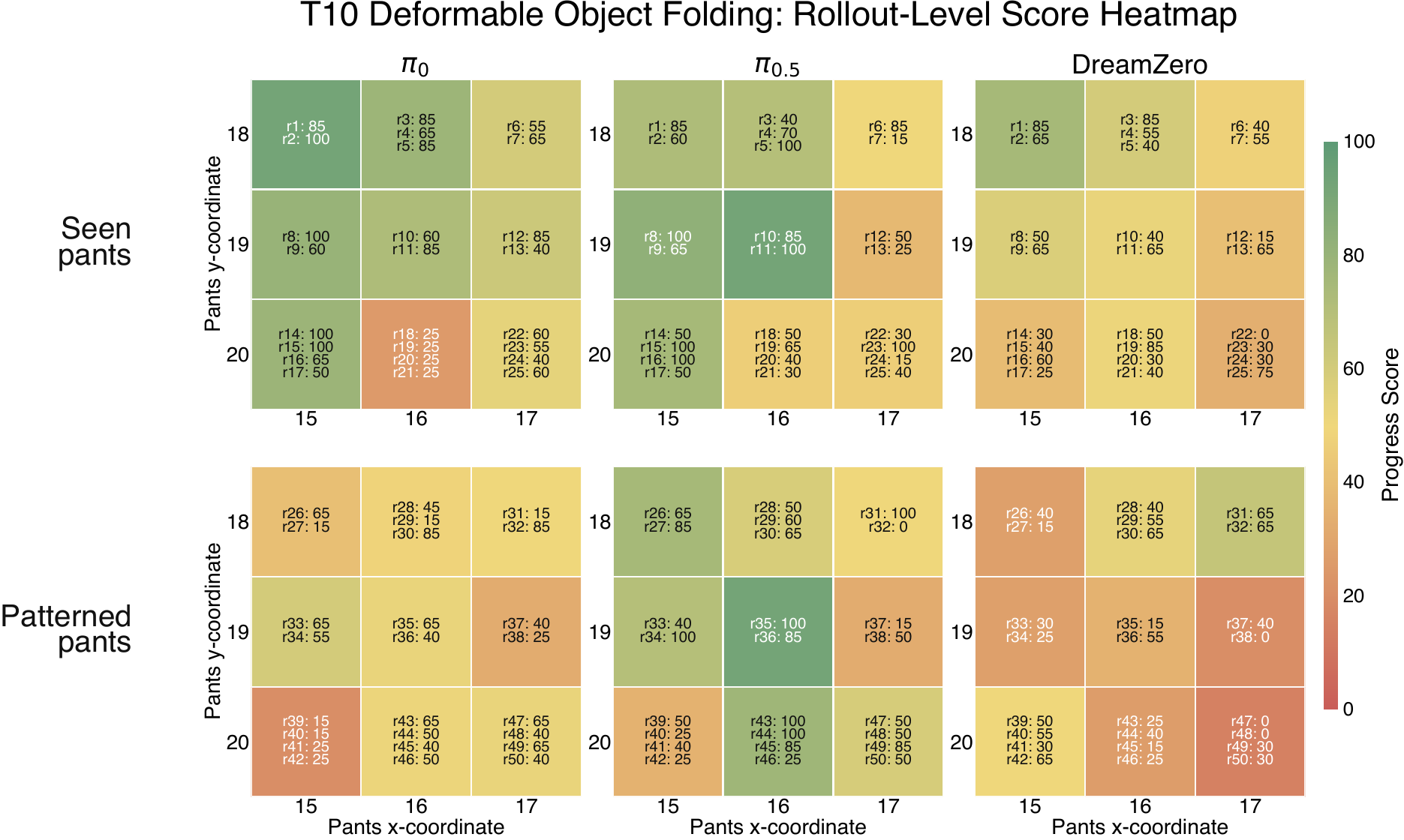}
    \caption{Task 10, Deformable Object Folding. Heatmap cells are physical pants-placement coordinates; entries report rollout ID and Progress Score.}
    \label{fig:bimanual_heatmap_t10}
\end{figure}

Figures~\ref{fig:bimanual_heatmap_t5}--\ref{fig:bimanual_heatmap_t10} visualize rollout-level progress scores for the six bimanual tasks.
For the grid-based tasks, the heatmap cells correspond to the physical reset coordinates used in the real-world evaluation rather than to abstract seen/unseen bins.
Panel labels and coordinate axes are task-specific: they denote basket, box, target-mat, mahjong-placement, long-horizon layout, or pants-placement conditions.
Tasks 5, 6, and 8 use $5\times5$ physical layouts, Task 10 uses the $3\times3$ pants-placement layout, and Tasks 7 and 9 use task-specific nonuniform layouts for the turntable-speed and long-horizon placement conditions.
Each cell label reports the rollout ID and its Progress Score; multiple lines inside one cell indicate repeated rollouts at the same physical or condition cell.

\subsection{Representative Task Process Analysis}
\label{app:t5_process_analysis}

\begin{figure}[H]
    \centering
    \includegraphics[max width=0.90\linewidth]{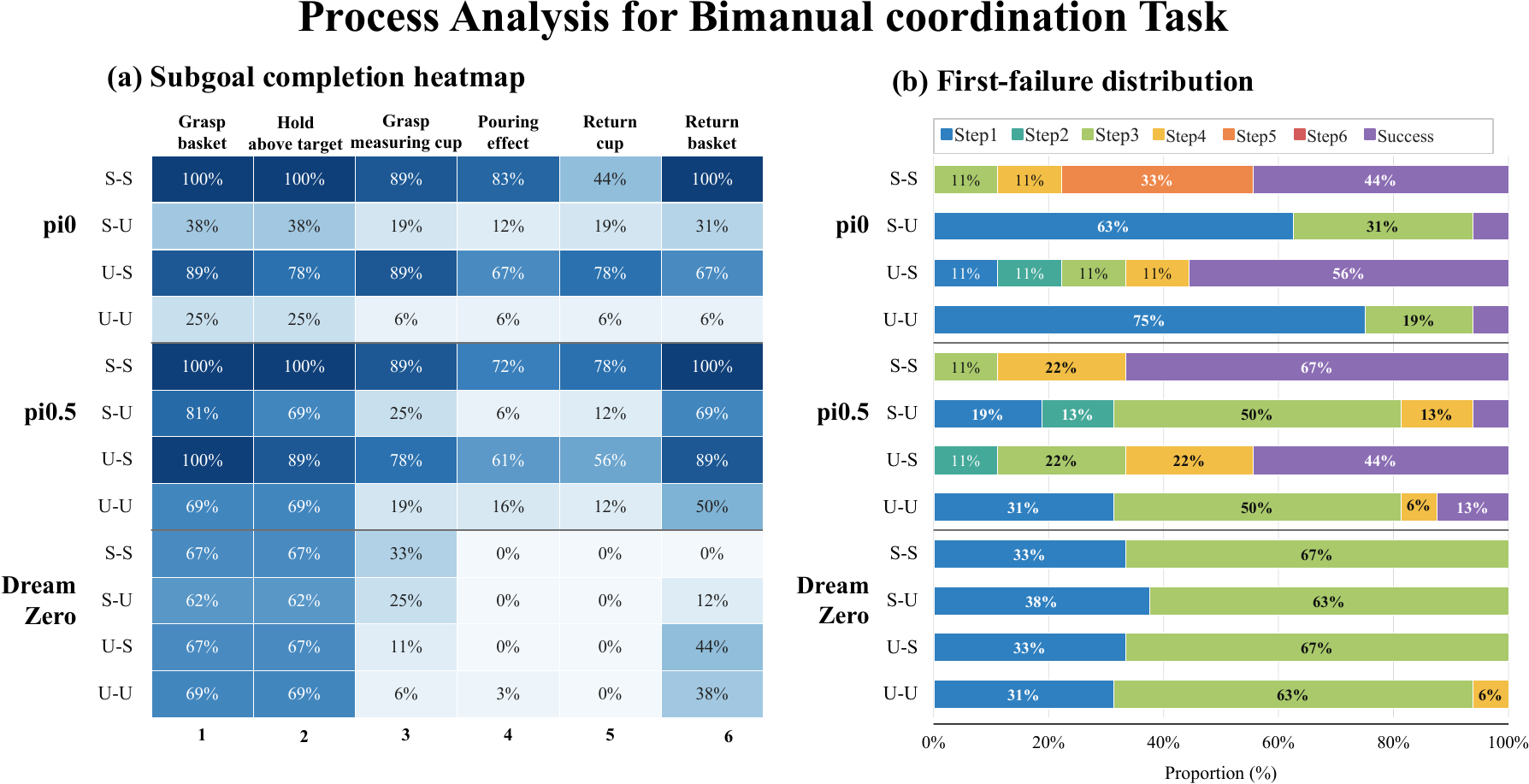}
    \caption{\textbf{Process analysis for Task 5, Bimanual Material Pouring.}
    Left: subgoal completion rates for each model and evaluation cell.
    Right: distribution of the first failed subgoal, with full completion shown as success.}
    \label{fig:t5_process_analysis}
\end{figure}

\subsection{Task 1: Sequential Object Stacking}

\paragraph{Goal.}
The robot stacks three plastic baskets in the designated target region, forming a stable three-layer structure.

\paragraph{Generalization variables.}
Seen episodes vary basket color, texture, position, and orientation within the training object set.
Unseen-object episodes use held-out basket appearances, while layout variation changes the initial basket positions within the reachable tabletop region.
The target mat remains fixed as the placement reference.

\paragraph{Evaluation protocol.}
The rollout is scored over three basket operations.
Basket 1 contributes 30 points, Basket 2 contributes 35 points, and Basket 3 contributes 35 points.
Each basket receives grasp credit only when it is lifted successfully, and placement or stacking credit only when it remains validly positioned in the final stack.

\paragraph{Full success criterion.}
A rollout is counted as a full success only if all three baskets receive full credit and the final stack is upright, stable, aligned, and free of major protrusion.

\paragraph{Scoring notes.}
If a basket is grasped but dropped before a valid placement or stacking event, only the corresponding grasp credit is awarded.
If a later action causes visible protrusion, large misalignment, or collapse, the affected placement or stacking credit is not awarded.
The final state must satisfy the stack stability requirement for full success.

\subsection{Task 2: Articulated Container Manipulation}

\paragraph{Goal.}
The robot opens a lidded box, inserts a garbage bag, and closes the lid.

\paragraph{Generalization variables.}
Seen episodes vary box pose, bag pose, and seen bag appearances.
Unseen-object episodes use held-out garbage-bag colors or appearance variants, while layout variation changes object positions and orientations within the task workspace.

\paragraph{Evaluation protocol.}
The rollout is scored over box opening, bag handling, and box closing.
The current rubric assigns 30 points to valid opening, 40 points to bag handling, and 30 points to valid closing.
Bag handling is tiered according to grasping, partial insertion, and full insertion retained after closing.

\paragraph{Full success criterion.}
A rollout is counted as a full success only if the lid is opened to a valid operating angle, the bag is fully inserted and retained, and the lid is closed flush without ejecting the bag or tipping the box.

\paragraph{Scoring notes.}
Slight lid motion that does not reach a valid operating angle does not count as opening.
If the bag remains protruding from the box opening, only partial-insertion credit is awarded.
Closing credit is not awarded if the lid rebounds, leaves a visible gap, ejects the bag, or causes the box to tip over.

\subsection{Task 3: Tool-Mediated Stamping}

\paragraph{Goal.}
The robot grasps a stamp, inks it on the ink pad, marks the target paper, and returns the stamp.

\paragraph{Generalization variables.}
Seen episodes vary stamp appearance, ink-pad state, paper color, and paper pose within the training combinations.
Unseen-object episodes use held-out stamp-paper or ink-paper combinations.
Layout variation changes the relative positions of the stamp, ink pad, and target paper.

\paragraph{Evaluation protocol.}
The rollout is scored over stamp grasping, inking, stamping, and stamp return.
The current rubric assigns 15 points to stamp grasping, 20 points to inking, 40 points to stamping, and 25 points to stamp return.
Stamping and return stages are tiered by mark quality and final stamp stability.

\paragraph{Full success criterion.}
A rollout is counted as a full success only if the stamp is grasped, inked, applied clearly within the target region, and returned upright and stable without tipping the ink pad or severely displacing the paper.

\paragraph{Scoring notes.}
Superficial contact with the ink pad does not count as successful inking; the stamp face must make effective contact.
No stamping credit is awarded if no visible mark is produced or if the mark falls entirely outside the target region.
Return credit depends on the stamp's final stable pose.

\subsection{Task 4: Precision Slot Insertion}

\paragraph{Goal.}
The robot places two remote controls into the corresponding slots of a remote-control holder.

\paragraph{Generalization variables.}
Seen episodes vary seen remote models, holder colors, and initial remote positions.
Unseen-object episodes use held-out remote models, holder colors, or remote-holder appearance combinations.
Layout variation changes the initial placement and orientation of the remotes relative to the holder.

\paragraph{Evaluation protocol.}
The rollout is scored over two remote-control sub-scores.
Each remote contributes 20 points for successful grasping and 30 points for stable slot placement, for a total of 100 points.
Credit is determined by the final stable state of each remote.

\paragraph{Full success criterion.}
A rollout is counted as a full success only if both remote controls are grasped, correctly inserted into their holder slots, and remain stable at rollout termination.

\paragraph{Scoring notes.}
If a remote is grasped but dropped before a valid slot placement, only the grasp credit for that remote is awarded.
If a remote slides out, tips over, is placed into the wrong region, or is displaced by a later action, its placement credit is not awarded.

\subsection{Task 5: Bimanual Material Pouring}

\paragraph{Goal.}
The robot transfers beans by using one arm to stabilize the receiving basket and the other arm to manipulate the measuring cup.
The task is completed when the beans are poured into the receiving basket and the relevant objects are returned to their target regions.

\paragraph{Generalization variables.}
The target mat is fixed as the central reference.
The basket and measuring cup are offset relative to the target mat on a grid, so larger offsets induce stronger relative layout changes rather than a simple translation of the whole scene.
Unseen episodes introduce held-out basket-bean combinations or held-out basket--measuring-cup layouts.

\paragraph{Evaluation protocol.}
The rollout is scored over bimanual stabilization, measuring-cup grasping, pouring, material containment, and object-return stages.
The maximum rollout budget is selected after non-scored timing runs and should allow a complete execution with one minor grasp reattempt.

\paragraph{Full success criterion.}
A rollout is counted as a full success only if the receiving basket remains stable, the measuring cup is manipulated without unsafe collision, the beans are transferred into the basket, and the final object states remain stable in the target regions.

\paragraph{Scoring notes.}
Credit is assigned from the final stable state.
If the cup is grasped but the beans are not poured into the basket, only the completed grasp or transport stages receive credit.
If the basket tips over, the beans are spilled outside the valid region, or either object ends in an invalid target region, the corresponding transfer or placement credit is not awarded.

\subsection{Task 6: Bimanual Packing and Transport}

\paragraph{Goal.}
The robot packs tabletop objects such as a tissue roll and a ball into a storage box, then transports the loaded box to the target area using both arms.

\paragraph{Generalization variables.}
Because the storage box is the primary interacted object during transport, layout generalization is defined mainly through box position.
Seen episodes use center-grid box positions, while unseen-layout episodes use outer-grid box positions.
The auxiliary objects retain fixed nominal centers so that the layout shift primarily tests bimanual transport and box handling.

\paragraph{Evaluation protocol.}
The rollout is scored over object packing, box grasping, bimanual lifting, coordinated transport, and final placement.
The task requires the two arms to maintain box stability while moving the packed objects to the target area.

\paragraph{Full success criterion.}
A rollout is counted as a full success only if the required objects are packed into the box, the box is transported with both arms without losing its contents, and the final box pose is stable inside the target region.

\paragraph{Scoring notes.}
Packing credit is not awarded for objects left outside the box at termination.
Transport credit is not awarded if the box is dragged into an invalid pose, dropped, tipped, or moved without retaining the packed contents.
Final-placement credit requires the box to remain stable in the target area.

\subsection{Task 7: Dynamic Pick-and-Place}

\paragraph{Goal.}
The robot grasps a can from a rotating turntable and places it on the target mat.

\paragraph{Generalization variables.}
Seen episodes use a medium turntable speed of 15 degrees per second with seen object appearances.
Unseen episodes vary the dynamic condition using lower and higher turntable speeds, including 8 degrees per second and 30 degrees per second, and also include held-out object appearances under the medium-speed setting.
This task does not define an unseen-position split; the non-object generalization axis is speed.

\paragraph{Evaluation protocol.}
The rollout is scored over dynamic target approach, successful grasp, lifting from the turntable, transport, and stable placement on the target mat.
The task evaluates timing-sensitive perception and control because the grasp target moves continuously during execution.

\paragraph{Full success criterion.}
A rollout is counted as a full success only if the can is grasped from the rotating turntable, transported without dropping, and placed stably on the target mat.

\paragraph{Scoring notes.}
Contact with the can without a stable lift does not receive grasp credit.
Placement credit is not awarded if the can is placed outside the target mat, falls over after placement, or is knocked away by a later motion.
Failures caused by mistimed approach or collision with the turntable are reflected in the dynamic-grasp stage.

\subsection{Task 8: Category Sorting and Placement}

\paragraph{Goal.}
The robot sorts four mahjong tiles into the corresponding baskets according to their category or color.

\paragraph{Generalization variables.}
Seen episodes use the training mahjong category combinations and basket arrangement.
Factor-A held-out episodes use unseen mahjong category combinations rather than novel physical object instances.
Factor-B held-out episodes change the left-right category arrangement or the order of target baskets while preserving the same high-level sorting rule.

\paragraph{Evaluation protocol.}
The rollout is scored over tile identification, tile grasping, transport, and placement into the matching basket.
Each tile contributes partial credit only when its final basket assignment is correct and stable.

\paragraph{Full success criterion.}
A rollout is counted as a full success only if all four tiles are placed into their matching baskets and remain in valid final positions.

\paragraph{Scoring notes.}
No placement credit is awarded for a tile placed into the wrong category basket.
If a tile is grasped and transported but dropped outside all valid baskets, only the completed grasp or transport stage receives credit.
If a later action displaces a previously correct tile into an invalid region, the final placement credit for that tile is not awarded.

\subsection{Task 9: Long-Horizon Rearrangement}

\paragraph{Goal.}
The robot rearranges multiple kitchen objects across several target regions: cups are placed on the upper rack, cans on the lower rack, and bread on the plate.

\paragraph{Generalization variables.}
Seen episodes use training object appearances such as seen cup and plate variants.
Unseen-object episodes use held-out cup, plate, or related object appearances.
Layout variation separates middle-cross placements from corner placements in the $3\times3$ placement region.
The task also stresses long-horizon generalization because correct completion depends on maintaining the ordering and target assignment of several object classes.

\paragraph{Evaluation protocol.}
The rollout is scored over repeated grasp, transport, and placement stages for cups, cans, and bread.
Cups must be placed on the upper rack, cans on the lower rack, and bread on the plate.
The task allows a minor reattempt within the maximum rollout budget, but credit is based on the final stable state.

\paragraph{Full success criterion.}
A rollout is counted as a full success only if all required objects are placed in their correct target regions, at the correct rack level when applicable, and remain stable at termination.

\paragraph{Scoring notes.}
If an object is placed on the wrong rack level, only grasp or transport credit is awarded for that object.
If an object is placed too close to an edge and later falls or becomes visibly unstable, placement credit is not awarded.
Bread placement receives full credit only when the main body of the bread remains stably inside the plate.

\subsection{Task 10: Deformable Object Folding}

\paragraph{Goal.}
The robot folds a pair of pants from an unfolded or semi-unfolded tabletop state into a compact and regular final configuration.

\paragraph{Generalization variables.}
Seen-object episodes use green and blue pants, while unseen-object episodes use patterned pants.
Seen-position episodes use the upper two rows of the $3\times3$ placement grid, while unseen-position episodes use the lower row.
The 50-rollout evaluation list contains 13 seen-object/seen-position rollouts, 12 seen-object/unseen-position rollouts, 13 unseen-object/seen-position rollouts, and 12 unseen-object/unseen-position rollouts.
The task is designed to test deformable-object handling, bimanual coordination, long-horizon execution, and layout generalization.

\paragraph{Evaluation protocol.}
The rollout is scored over cloth grasping, spreading or alignment, coordinated folding, final compactness, and final regularity.
Unlike rigid-object tasks, the final state is judged by geometric regularity and stability of the folded cloth rather than by a single rigid pose.

\paragraph{Full success criterion.}
A rollout is counted as a full success only if the pants are folded into a compact, stable, and visually regular final state, with no large unfolded regions left outside the intended folded footprint.

\paragraph{Scoring notes.}
Partial credit is awarded for meaningful progress such as successful cloth grasping, partial alignment, or a partially completed fold.
Credit is reduced when the cloth is twisted, dragged far from the workspace, left mostly unfolded, or folded into an unstable or irregular configuration.

\section{Additional Result Diagnostics}
\label{app:additional_result_diagnostics}

This appendix expands the compact discussion in Section~\ref{sec:experiments} with task-level diagnostics from the same evaluation records reported in Table~\ref{tab:main_results}.
Figure~\ref{fig:result_summary_plots} re-plots the Overall Score columns as a capability-wise summary and visualizes the main generalization trends, while Appendices~\ref{app:single_arm_heatmaps} and~\ref{app:bimanual_heatmaps} expand the same records to rollout-level scores and physical reset layouts.
No additional metrics are introduced here; the goal is to contextualize the aggregate trends reported in the main text.

\paragraph{Task-level model profiles.}
Although $\pi_{0.5}$ achieves the highest average Overall Score, its advantage is not uniform across all tasks.
It ranks first on T4, T5, T7, T8, T9, and T10, covering precision insertion, bimanual pouring, dynamic pick-and-place, semantic sorting, long-horizon rearrangement, and deformable-object folding.
By contrast, $\pi_0$ is strongest on the two single-arm placement tasks T1 and T2, leading on both Overall Score (75.00 and 86.50) and FSR (34
DreamZero trails on average but is sharply differentiated: it achieves the best Overall Score on T3 (34.30) and T6 (72.50), and its T6 result includes the highest FSR among all model-task pairs at 52
These task-level differences indicate that the three policies fail in different ways rather than following a single global ranking on every manipulation setting.

\paragraph{DreamZero's mixed profile.}
DreamZero's per-task profile points to both useful temporal context and deployment limitations.
Its advantage on T3 is consistent with the need to infer latent task stage from observation history, especially when a single wrist-view frame is ambiguous.
However, its weak results on T5, T7, and T9 are unlikely to have a single cause.
For T7, the latency profile in Appendix~\ref{app:inference_latency} suggests limited responsiveness to moving targets, which is particularly harmful in closed-loop dynamic manipulation.
For geometry-sensitive near-field tasks, we additionally hypothesize that DreamZero's world-model pretraining may not fully transfer to the fisheye wrist-view observation interface used in UMI-Bench.
This domain-transfer explanation remains a hypothesis and would require controlled ablations to isolate from task difficulty, action chunking, and inference speed.

\paragraph{Hardest tasks.}
T3 and T9 are the hardest tasks under the current protocol, with all three models obtaining 0\% FSR.
For T3, the main challenge is multi-stage progress estimation under an ambiguous single wrist view: policies often complete intermediate steps but fail to maintain the correct stage estimate until final completion.
For T9, errors accumulate across a long rearrangement sequence, and small deviations in early placements can make the final layout fail the tight spatial tolerance.
These two tasks therefore stress different failure modes: T3 emphasizes latent stage inference, whereas T9 emphasizes long-horizon error accumulation and final-placement precision.

\paragraph{Spatial-shift sensitivity.}
The main text reports that performance degrades more under spatial shifts than under object or appearance shifts.
This pattern is especially clear on T5, T6, and T8, where Factor-B drops dominate the corresponding Factor-A drops.
For example, $\pi_0$ on T5 falls from 86.67 in the Seen/Seen condition to 22.81 under positional shift, as shown in Figure~\ref{fig:t5_process_analysis}.
Such failures suggest that current policies partly rely on demonstration-aligned motion priors instead of consistently re-localizing task-relevant geometry at deployment.
Because the standardized UMI protocol maps the central collection grid directly onto the Seen/Seen condition, that column can be interpreted as a training-reproduction baseline.
The mean Seen/Seen score of 59.62 confirms that policies retain much of their capability when training and evaluation are aligned, whereas the combined-shift mean of 40.19 quantifies the loss once object or spatial distributions move beyond the recorded coverage.

\paragraph{Progress--success mismatch on T1.}
Task 1 illustrates why Progress Score and FSR are reported separately.
$\pi_0$ achieves 34
Thus, similar partial progress can mask different completion reliability: $\pi_{0.5}$ often reaches intermediate stacking stages but less often satisfies the final stability and alignment criterion.
This mismatch is important for real-world manipulation because partial task progress does not necessarily imply deployable task completion.

\paragraph{Balanced perturbation sensitivity on T10.}
T10 differs from the grid-layout tasks that drive the main Factor-B trend.
Its model-averaged Factor-A and Factor-B drops from Seen/Seen are nearly equal, at 16.28 and 16.17 points, while the combined-shift drop is 23.11 points.
We therefore treat T10 as a probe of general deformable-object execution robustness rather than specifically spatial robustness.
Changing pants appearance and changing the initial cloth placement both perturb grasp selection, cloth alignment, and subsequent fold execution, making the task sensitive to both visual and geometric variation.

\section{Tabular Scoring Rubrics}
\label{app:scoring_rubrics}

This appendix gives the task-level point allocations used to compute the 0--100 progress score.
Unless otherwise stated, credit is assigned from the final stable physical state at rollout termination.
For tiered stages, the highest satisfied tier is used rather than summing tiers within the same stage.
A rollout receives full-success credit only when the corresponding rubric total is 100 and the final physical state is stable.

\begin{figure}[H]
    \centering
    \includegraphics[max width=\linewidth]{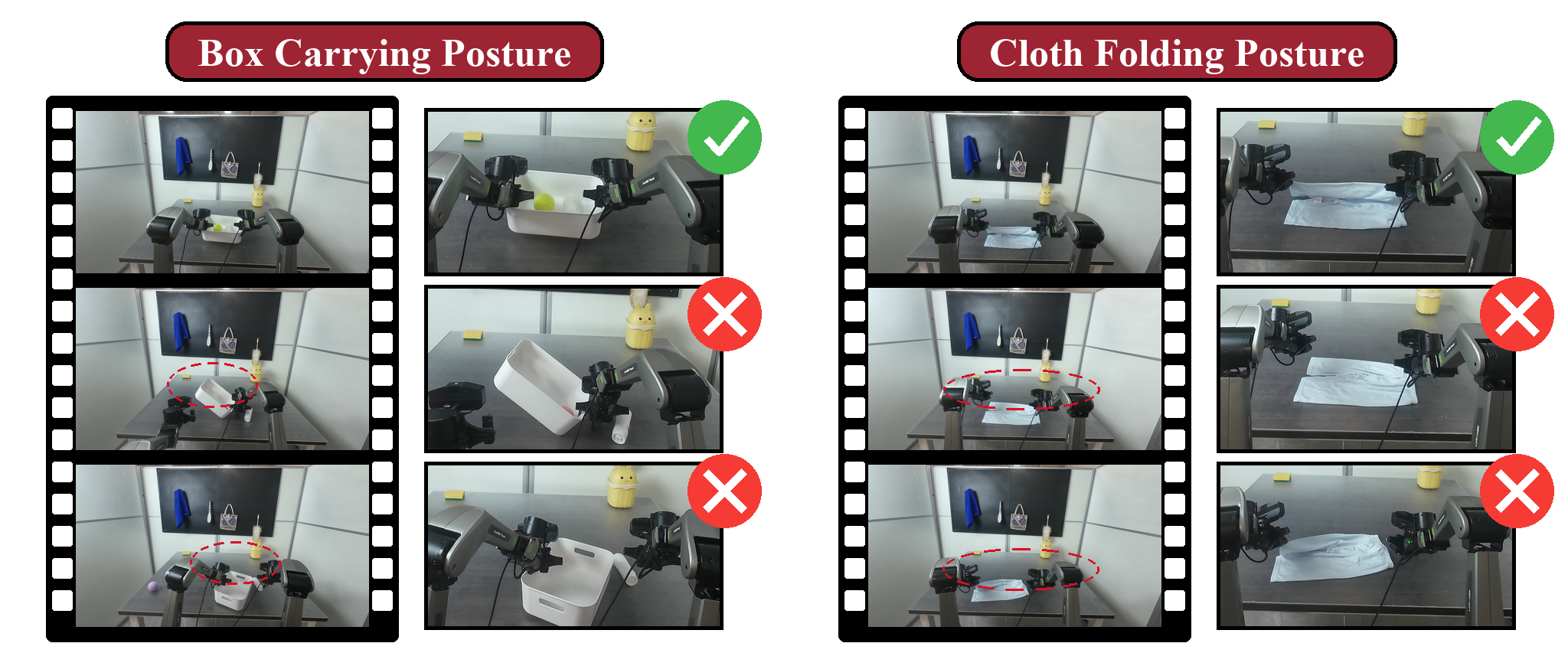}
    \caption{\textbf{Representative final-state scoring examples.}
    Third-person views illustrate valid and invalid final postures for box carrying and cloth folding.
    The same final-state convention is used when assigning full-success labels and partial Progress Scores.}
    \label{fig:posture_success_failure}
\end{figure}

\begin{table}[H]
    \centering
    \small
    \setlength{\tabcolsep}{4pt}
    \renewcommand{\arraystretch}{1.18}
    \caption{Scoring rubric for Task 1: Sequential Object Stacking.}
    \label{tab:task1_scoring_rubric}
    \begin{tabularx}{\linewidth}{>{\raggedright\arraybackslash}p{0.23\linewidth}>{\centering\arraybackslash}p{0.10\linewidth}>{\raggedright\arraybackslash}X}
        \toprule
        \textbf{Stage} & \textbf{Points} & \textbf{Criterion} \\
        \midrule
        Basket 1 & 30 & +10 grasp/lift; +20 placed on the target mat. \\
        Basket 2 & 35 & +10 grasp/lift; +25 stacked on Basket 1, aligned and stable. \\
        Basket 3 & 35 & +10 grasp/lift; +25 stacked on Basket 2, aligned and stable. \\
        \midrule
        Full success & 100 & All baskets receive full credit; final stack is upright, stable, and has no major protrusion. \\
        \bottomrule
    \end{tabularx}
\end{table}

\begin{table}[H]
    \centering
    \small
    \setlength{\tabcolsep}{4pt}
    \renewcommand{\arraystretch}{1.18}
    \caption{Scoring rubric for Task 2: Articulated Container Manipulation.}
    \label{tab:task2_scoring_rubric}
    \begin{tabularx}{\linewidth}{>{\raggedright\arraybackslash}p{0.23\linewidth}>{\centering\arraybackslash}p{0.10\linewidth}>{\raggedright\arraybackslash}X}
        \toprule
        \textbf{Stage} & \textbf{Points} & \textbf{Criterion} \\
        \midrule
        Box opening & 30 & Lid reaches a valid open angle; box remains upright. \\
        Garbage-bag handling & 40 & Tiered: +15 grasp/lift; +25 partial insertion; +40 full insertion retained after closing. \\
        Box closing & 30 & Lid closes flush with no visible gap, no bag ejection, and no box tipping. \\
        \midrule
        Full success & 100 & Valid opening, full bag insertion, flush closing, no bag ejection or box tipping. \\
        \bottomrule
    \end{tabularx}
\end{table}

\begin{table}[H]
    \centering
    \small
    \setlength{\tabcolsep}{4pt}
    \renewcommand{\arraystretch}{1.18}
    \caption{Scoring rubric for Task 3: Tool-Mediated Stamping.}
    \label{tab:task3_scoring_rubric}
    \begin{tabularx}{\linewidth}{>{\raggedright\arraybackslash}p{0.23\linewidth}>{\centering\arraybackslash}p{0.10\linewidth}>{\raggedright\arraybackslash}X}
        \toprule
        \textbf{Stage} & \textbf{Points} & \textbf{Criterion} \\
        \midrule
        Stamp grasp & 15 & Stamp grasped and lifted from the tray. \\
        Inking & 20 & Stamp face makes effective ink-pad contact. \\
        Stamping & 40 & Tiered: +20 any visible in-target mark; +40 clear, discernible in-target mark. \\
        Stamp return & 25 & Tiered: +10 returned to tray region; +25 returned upright and stable. \\
        \midrule
        Full success & 100 & All stages receive full credit; clear in-target mark; no ink-pad tip or severe paper displacement. \\
        \bottomrule
    \end{tabularx}
\end{table}

\begin{table}[H]
    \centering
    \small
    \setlength{\tabcolsep}{4pt}
    \renewcommand{\arraystretch}{1.18}
    \caption{Scoring rubric for Task 4: Precision Slot Insertion.}
    \label{tab:task4_scoring_rubric}
    \begin{tabularx}{\linewidth}{>{\raggedright\arraybackslash}p{0.23\linewidth}>{\centering\arraybackslash}p{0.10\linewidth}>{\raggedright\arraybackslash}X}
        \toprule
        \textbf{Stage} & \textbf{Points} & \textbf{Criterion} \\
        \midrule
        Remote 1 grasp & 20 & Remote 1 grasped and lifted off the table. \\
        Remote 1 placement & 30 & Remote 1 placed into the holder slot and remains stable. \\
        Remote 2 grasp & 20 & Remote 2 grasped and lifted stably off the table. \\
        Remote 2 placement & 30 & Remote 2 placed into the holder slot and remains stable. \\
        \midrule
        Full success & 100 & Both remotes receive full credit; both are correctly slotted and stable. \\
        \bottomrule
    \end{tabularx}
\end{table}

\begin{table}[H]
    \centering
    \small
    \setlength{\tabcolsep}{4pt}
    \renewcommand{\arraystretch}{1.18}
    \caption{Scoring rubric for Task 5: Bimanual Material Pouring.}
    \label{tab:task5_scoring_rubric}
    \begin{tabularx}{\linewidth}{>{\raggedright\arraybackslash}p{0.23\linewidth}>{\centering\arraybackslash}p{0.10\linewidth}>{\raggedright\arraybackslash}X}
        \toprule
        \textbf{Stage} & \textbf{Points} & \textbf{Criterion} \\
        \midrule
        Basket grasp & 10 & Left arm grasps the empty basket and lifts it stably from the table. \\
        Basket stabilization & 15 & Basket is moved to the target-mat region and held stable during pouring. \\
        Measuring-cup grasp & 10 & Right arm grasps the bean-filled measuring cup and lifts it stably from the table. \\
        Bean transfer & 40 & Tiered: +20 pouring action completed with partial bean transfer or visible spill; +40 most beans enter the basket with the main transfer completed. \\
        Cup return & 10 & Measuring cup is returned to the designated final region. \\
        Basket return & 15 & Basket containing the transferred beans is placed back stably in the target-mat region. \\
        \midrule
        Full success & 100 & Basket is stabilized, beans are transferred into it, both objects are returned, and no severe spill, tip, or invalid final pose occurs. \\
        \bottomrule
    \end{tabularx}
\end{table}

\begin{table}[H]
    \centering
    \small
    \setlength{\tabcolsep}{4pt}
    \renewcommand{\arraystretch}{1.18}
    \caption{Scoring rubric for Task 6: Bimanual Packing and Transport.}
    \label{tab:task6_scoring_rubric}
    \begin{tabularx}{\linewidth}{>{\raggedright\arraybackslash}p{0.23\linewidth}>{\centering\arraybackslash}p{0.10\linewidth}>{\raggedright\arraybackslash}X}
        \toprule
        \textbf{Stage} & \textbf{Points} & \textbf{Criterion} \\
        \midrule
        Ball grasp & 10 & Left arm grasps the ball and lifts it stably from the table. \\
        Ball packing & 10 & Ball is placed inside the storage box and does not bounce or remain outside. \\
        Tissue-roll grasp & 10 & Right arm grasps the tissue roll and lifts it stably from the table. \\
        Tissue-roll packing & 10 & Tissue roll is placed inside the storage box and does not roll out or remain outside. \\
        Left handle grasp & 5 & Left arm makes effective contact with and holds the left handle. \\
        Right handle grasp & 5 & Right arm makes effective contact with and holds the right handle. \\
        Bimanual lift & 20 & Both arms jointly lift the storage box without obvious one-sided dragging, tipping, or severe tilt. \\
        Bimanual transport & 30 & Tiered: +15 valid transport distance is at most 15 cm; +30 valid transport distance exceeds 15 cm. \\
        \midrule
        Full success & 100 & Both objects are packed, the loaded box is lifted with both handles, transported beyond the full-distance tier, and remains stable. \\
        \bottomrule
    \end{tabularx}
\end{table}

\begin{table}[H]
    \centering
    \small
    \setlength{\tabcolsep}{4pt}
    \renewcommand{\arraystretch}{1.18}
    \caption{Scoring rubric for Task 7: Dynamic Pick-and-Place.}
    \label{tab:task7_scoring_rubric}
    \begin{tabularx}{\linewidth}{>{\raggedright\arraybackslash}p{0.23\linewidth}>{\centering\arraybackslash}p{0.10\linewidth}>{\raggedright\arraybackslash}X}
        \toprule
        \textbf{Stage} & \textbf{Points} & \textbf{Criterion} \\
        \midrule
        Right dynamic grasp & 30 & Tiered: +15 can is briefly held but unstable; +30 can is grasped from the rotating turntable and lifted away stably. \\
        Right target placement & 20 & Tiered: +10 can partially contacts the right target mat but final pose is imperfect; +20 can is placed stably on the right target mat. \\
        Left dynamic grasp & 30 & Tiered: +15 can is briefly held but unstable; +30 can is grasped from the rotating turntable and lifted away stably. \\
        Left target placement & 20 & Tiered: +10 can partially contacts the left target mat but final pose is imperfect; +20 can is placed stably on the left target mat. \\
        \midrule
        Full success & 100 & Both cans are grasped from the moving turntable and placed stably on their corresponding target mats. \\
        \bottomrule
    \end{tabularx}
\end{table}

\begin{table}[H]
    \centering
    \small
    \setlength{\tabcolsep}{4pt}
    \renewcommand{\arraystretch}{1.18}
    \caption{Scoring rubric for Task 8: Category Sorting and Placement.}
    \label{tab:task8_scoring_rubric}
    \begin{tabularx}{\linewidth}{>{\raggedright\arraybackslash}p{0.23\linewidth}>{\centering\arraybackslash}p{0.10\linewidth}>{\raggedright\arraybackslash}X}
        \toprule
        \textbf{Stage} & \textbf{Points} & \textbf{Criterion} \\
        \midrule
        Left tile 1 & 25 & +5 grasp/lift; +20 stable placement in the correct category basket. Placement credit is capped at +5 if the tile is stably placed in a wrong-category basket. \\
        Left tile 2 & 25 & +5 grasp/lift; +20 stable placement in the correct category basket. Placement credit is capped at +5 if the tile is stably placed in a wrong-category basket. \\
        Right tile 1 & 25 & +5 grasp/lift; +20 stable placement in the correct category basket. Placement credit is capped at +5 if the tile is stably placed in a wrong-category basket. \\
        Right tile 2 & 25 & +5 grasp/lift; +20 stable placement in the correct category basket. Placement credit is capped at +5 if the tile is stably placed in a wrong-category basket. \\
        \midrule
        Full success & 100 & All four tiles are grasped and placed stably in their matching category baskets. \\
        \bottomrule
    \end{tabularx}
\end{table}

\begin{table}[H]
    \centering
    \small
    \setlength{\tabcolsep}{4pt}
    \renewcommand{\arraystretch}{1.18}
    \caption{Scoring rubric for Task 9: Long-Horizon Rearrangement.}
    \label{tab:task9_scoring_rubric}
    \begin{tabularx}{\linewidth}{>{\raggedright\arraybackslash}p{0.23\linewidth}>{\centering\arraybackslash}p{0.10\linewidth}>{\raggedright\arraybackslash}X}
        \toprule
        \textbf{Stage} & \textbf{Points} & \textbf{Criterion} \\
        \midrule
        Cup 1 grasp & 10 & Cup 1 is grasped successfully. \\
        Cup 1 placement & 10 & Cup 1 is placed stably in the upper-left target region of the two-layer rack. \\
        Cup 2 grasp & 10 & Cup 2 is grasped successfully. \\
        Cup 2 placement & 10 & Cup 2 is placed stably in the upper-right target region of the two-layer rack. \\
        Can 1 grasp & 10 & Can 1 is grasped successfully. \\
        Can 1 placement & 10 & Can 1 is placed stably in the lower-rack target region without falling or obvious instability. \\
        Can 2 grasp & 10 & Can 2 is grasped successfully. \\
        Can 2 placement & 10 & Can 2 is placed stably in the lower-rack target region without falling or obvious instability. \\
        Bread grasp & 10 & Bread is grasped or supported without being visibly pushed away or dropped. \\
        Bread placement & 10 & Bread is placed on the plate; full credit requires its main body to remain stably inside the valid plate region. \\
        \midrule
        Full success & 100 & Both cups are on the upper rack, both cans are on the lower rack, bread is on the plate, and all objects remain stable. \\
        \bottomrule
    \end{tabularx}
\end{table}

\begin{table}[H]
    \centering
    \small
    \setlength{\tabcolsep}{4pt}
    \renewcommand{\arraystretch}{1.18}
    \caption{Scoring rubric for Task 10: Deformable Object Folding.}
    \label{tab:task10_scoring_rubric}
    \begin{tabularx}{\linewidth}{>{\raggedright\arraybackslash}p{0.23\linewidth}>{\centering\arraybackslash}p{0.10\linewidth}>{\raggedright\arraybackslash}X}
        \toprule
        \textbf{Stage} & \textbf{Points} & \textbf{Criterion} \\
        \midrule
        First left grasp & 15 & Left arm stably grasps a pant leg or pant-edge region. \\
        First right grasp & 15 & Right arm stably grasps a pant leg or pant-edge region. \\
        First fold & 30 & Tiered: +10 partial leg fold toward the pants body; +30 leg region is clearly folded over the pants body, forming the first fold. \\
        Second-fold grasp & 15 & After the first fold, left arm stably grasps a pant leg or the folded cloth body for the next fold. \\
        Second fold & 25 & Tiered: +10 partial second fold; +25 pants are further folded into a compact, stable, and visually regular final shape. \\
        \midrule
        Full success & 100 & All stages receive full credit; the final pants state is compact, stable, and regular with no large unfolded region outside the folded footprint. \\
        \bottomrule
    \end{tabularx}
\end{table}

\section{Evaluation Checklist}
\label{app:evaluation_checklist}

Table~\ref{tab:evaluation_checklist} summarizes the operator checks used to make each evaluation rollout auditable.

\begin{table}[H]
    \centering
    \caption{Evaluation checklist for real-world UMI-Bench rollouts.}
    \label{tab:evaluation_checklist}
    \footnotesize
    \begingroup
    \setlength{\tabcolsep}{4pt}
    \renewcommand{\arraystretch}{1.08}
    \begin{tabularx}{\linewidth}{>{\raggedright\arraybackslash}m{0.22\linewidth}>{\raggedright\arraybackslash}X}
        \toprule
        \textbf{Stage} & \textbf{Required checks} \\
        \midrule
        Before rollout & Verify tabletop setup, object set, lighting condition, wrist-camera stream, robot calibration, policy checkpoint, software version, logging directory, and scoring script. \\
        Scene reset & Match the reset image and scene JSON, including task ID, object IDs, appearance fields, tabletop positions, target regions, and split labels. \\
        During rollout & Confirm that the policy runs under the assigned checkpoint and that wrist video, robot state, action chunks, timestamps, and runtime logs are being recorded. \\
        After rollout & Check that the result package contains metadata, scene JSON, wrist video, trajectory logs, action logs, runtime logs, and score JSON. \\
        Scoring audit & Assign Progress Score and full-success labels from the recorded final state, then verify that the score JSON matches the task rubric and any failure notes. \\
        \bottomrule
    \end{tabularx}
    \endgroup
\end{table}

\end{document}